\documentclass[final,5p,times,numbers]{elsarticle}

\usepackage[utf8]{inputenc}
\usepackage[T1]{fontenc}
\usepackage{amsmath,amssymb,amsfonts,mathtools,amsthm}
\usepackage{booktabs}
\usepackage{multirow}
\usepackage{makecell,array,tabularx}
\usepackage[table]{xcolor}
\usepackage{graphicx}
\usepackage{subcaption}
\usepackage{pifont}
\usepackage{enumitem}
\usepackage{threeparttable}
\usepackage{rotating}

\usepackage{microtype}
\usepackage{url}
\usepackage{float}
\usepackage{placeins}
\usepackage{dblfloatfix}
\usepackage{listings}
\usepackage{caption}
\usepackage{algorithm}
\usepackage{algpseudocode}

\usepackage[colorlinks=true,linkcolor=blue,citecolor=blue,urlcolor=blue]{hyperref}
\usepackage[capitalize,noabbrev]{cleveref}
\graphicspath{{figure/}}
\DeclareGraphicsExtensions{.png}

\journal{Swarm and Evolutionary Computation}

\newcolumntype{L}[1]{>{\raggedright\arraybackslash}p{#1}}
\newcolumntype{C}[1]{>{\centering\arraybackslash}p{#1}}

\DeclareUnicodeCharacter{03C0}{\ensuremath{\pi}}
\DeclareUnicodeCharacter{2013}{-}
\DeclareUnicodeCharacter{2014}{,}
\DeclareUnicodeCharacter{2019}{'}
\algrenewcommand\algorithmicrequire{\textbf{Input:}}
\algrenewcommand\algorithmicensure{\textbf{Output:}}
\makeatletter
\providecommand{\theHALG@line}{\thealgorithm.\arabic{ALG@line}}
\makeatother


\begin{document}

\begin{frontmatter}

\title{GNBG-C: A Unified and Configurable Benchmark Generator for Constrained Continuous Optimization}

\author[first]{Ali Ghasemzadeh}
\ead{ali1383@hku.hk}
\author[first]{Erfan Babakhani}
\ead{erfan166@hku.hk}
\author[first]{Armin Khosravi}
\ead{arminkh@hku.hk}

\author[second]{Pariya Ghasemzadeh}
\ead{pariyaghasemzadeh101@gmail.com}

\author[first]{Yong-Hong~Kuo}
\ead{yhkuo@hku.hk}

\author[third,fourth,sixth]{Seyedali Mirjalili}
\author[fifth,sixth]{Amir H. Gandomi \corref{cor1}}
\ead{gandomi@uts.edu.au}
\cortext[cor1]{Corresponding author}

\address[first]{Department of Data and Systems Engineering, the University of Hong Kong, Hong Kong SAR, China}

\address[second]{Independent Researcher, Isfahan, Iran}

\address[third]{Centre for Artificial Intelligence Research and Optimisation, Torrens University Australia, Brisbane, Australia}

\address[fourth]{Faculty of Electrical Engineering and Computer Science, VSB Technical University of Ostrava, Ostrava, Czech Republic}

\address[fifth]{University of Technology Sydney, Sydney, Australia}
\address[sixth]{Obuda University, Budapest, Hungary}

\begin{abstract}
Many real-world optimization tasks are governed not only by what should be optimized, but also by what is allowed: a solution must first satisfy physical, operational, safety, budgetary, or structural requirements before its objective value becomes meaningful. Yet benchmark design for constrained continuous optimization often treats these restrictions as fixed properties of predefined test instances. Existing generators and benchmark suites provide essential foundations for reproducible comparison, but they either focus primarily on configurable objective landscapes or expose constraint properties within limited, predefined regimes, making it difficult to ask targeted questions about constraint-induced difficulty: whether an optimizer struggles because feasible regions are small, boundaries are restrictive, variables are tightly coupled, or constraint information is poorly scaled. We introduce GNBG-C, a Generalized Numerical Benchmark Generator for Constrained Continuous Optimization. The central idea is to treat the feasible region itself as a configurable object rather than as an inseparable attachment to a fixed test function. By generating controlled constraint regimes around the known GNBG global minimizer, the generator enables systematic studies in which feasibility pressure can be varied while preserving that point as a global minimizer of the constrained problem. This allows researchers to examine how optimizers discover feasibility, handle constraint pressure, and improve objective quality under diverse and controllable feasibility conditions. GNBG-C therefore provides a flexible and auditable framework for probing the strengths and weaknesses of constrained continuous optimizers under targeted forms of feasibility-side difficulty.

\end{abstract}


\end{frontmatter}

\section{Introduction}
\label{sec:introduction}

Many real-world optimization tasks are governed not only by what should be optimized, but also by what is allowed. In engineering design~\cite{holmberg2013stress}, operations, resource allocation~\cite{hartmann2010survey}, control~\cite{rawlings2017model}, and scientific computing~\cite{nocedal2006numerical}, a solution must often satisfy physical laws, safety requirements, budget limits, structural conditions, or operational restrictions before its objective value becomes meaningful. A design with excellent nominal performance may be unusable if it violates stress limits~\cite{holmberg2013stress}; a schedule with low cost may be infeasible if it exceeds resource availability~\cite{hartmann2010survey}; and a parameter setting with high predicted quality may be invalid if it violates system constraints~\cite{berkenkamp2023bayesian}. This makes constrained continuous optimization fundamentally different from unconstrained objective minimization: an optimizer must not only search for good objective values, but also discover, maintain, and exploit feasibility~\cite{michalewicz1996evolutionary,coello2002theoretical}.

This distinction has important consequences for benchmarking~\cite{hansen2021coco,hansen2009bbob}. In unconstrained optimization, the difficulty of a problem is largely associated with the objective landscape: modality, conditioning, separability, variable interaction, basin structure, and deceptiveness~\cite{hansen2009bbob,yazdani2023gnbg}. In constrained optimization, these objective-side properties remain important, but they are only part of the story~\cite{michalewicz1996evolutionary,coello2002theoretical}. The feasible region introduces an additional source of difficulty~\cite{michalewicz2000testcase}. Feasible points may occupy only a small portion of the search domain~\cite{michalewicz2000testcase}; constraint boundaries may be curved, restrictive, or highly nonconvex~\cite{michalewicz2000testcase,liang2006problem}; active constraints may shape the neighborhood of high-quality solutions~\cite{nocedal2006numerical}; variables may be coupled through dense constraint expressions~\cite{liang2006problem}; and numerical scaling may distort the apparent importance of different violations~\cite{runarsson2000stochastic,coello2002theoretical}. A constrained optimizer may therefore fail for several different reasons: it may search the objective poorly~\cite{hansen2009bbob,yazdani2023gnbg}, fail to locate feasible regions~\cite{michalewicz2000testcase,deb2000efficient}, reduce violation too slowly~\cite{coello2002theoretical,runarsson2000stochastic}, or lose objective progress while attempting to maintain feasibility~\cite{deb2000efficient,runarsson2000stochastic}.

\begin{figure*}[t]
    \centering
    \includegraphics[width=0.9\linewidth]{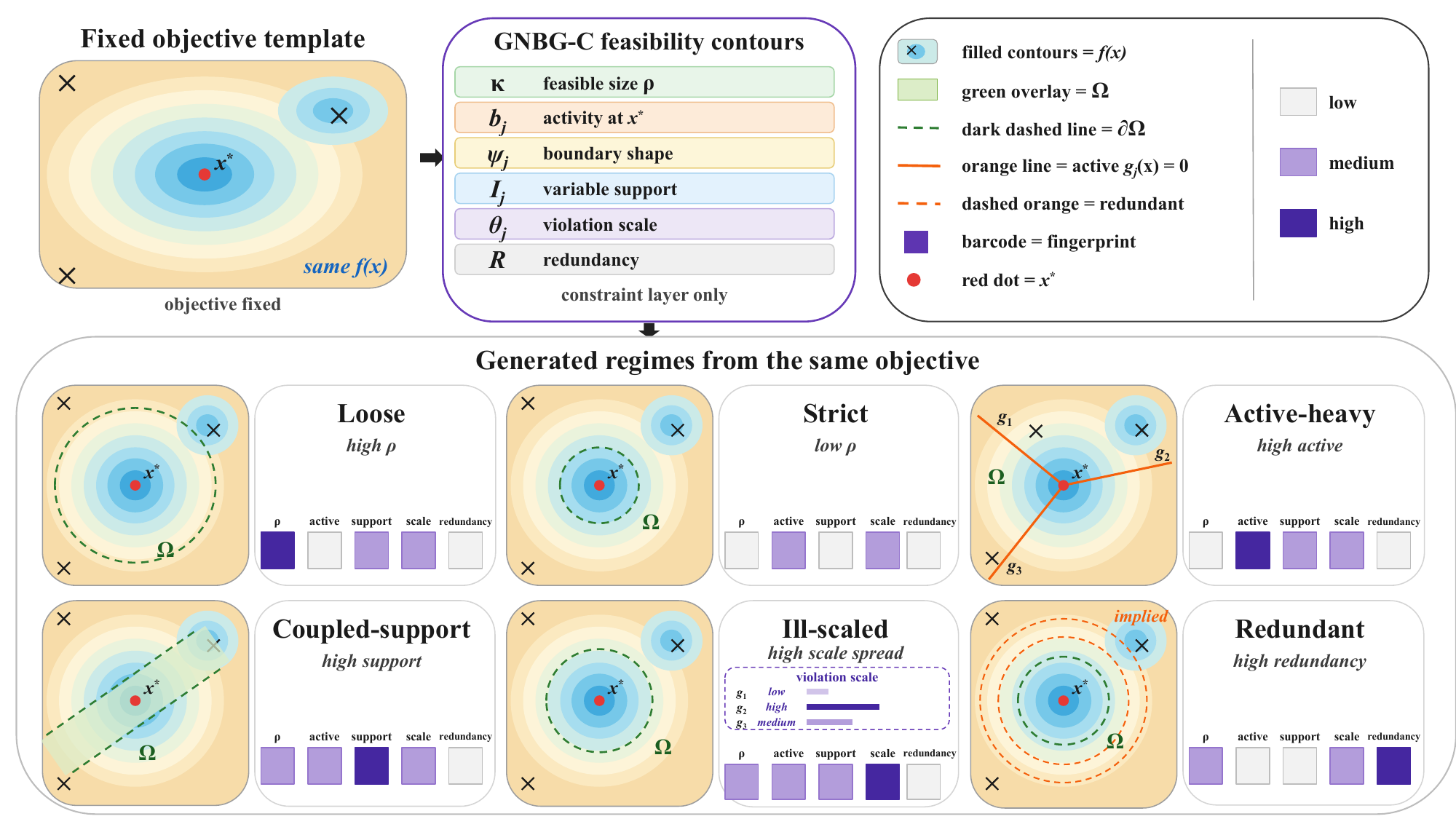}
    \caption{
    Core idea of GNBG-C. The objective landscape is kept fixed, while the generated constraint layer induces different feasible regions, such as loose, and strict regimes. This separates objective-landscape difficulty from feasible-region difficulty. The two-dimensional view is for interpretability; GNBG-C supports arbitrary user-specified dimensionality.
    }
    \label{fig:intro_objective_feasible_regions}
\end{figure*}

A benchmark for constrained continuous optimization should ideally help distinguish between these sources of behavior~\cite{michalewicz2000testcase,hellwig2019benchmarking}. Benchmarking is not only a matter of ranking algorithms by final objective value; it is also a tool for understanding the conditions under which different constraint-handling strategies succeed or fail~\cite{coello2002theoretical,hellwig2019benchmarking}. Shared benchmark suites and black-box benchmarking platforms provide the common ground required for reproducible comparison~\cite{hansen2021coco,bbobconstrained2024,hellwig2019benchmarking,liang2006problem}. Their strength is standardization: algorithms are evaluated on the same test problems, under shared protocols, with comparable performance indicators~\cite{hansen2021coco,bbobconstrained2024,liang2006problem}. This has made fixed suites and benchmarking platforms essential to the development and assessment of constrained optimizers~\cite{hellwig2019benchmarking,liang2006problem}.

However, common test instances do not always make the source of difficulty clear~\cite{hellwig2019benchmarking,michalewicz2000testcase,yazdani2023gnbg}. In a fixed constrained test instance, multiple sources of difficulty are usually embedded together~\cite{hellwig2019benchmarking,liang2006problem}. The objective landscape, feasible-region size, constraint geometry, active-set structure, variable coupling, coefficient scaling, and redundancy are all determined by the same predefined problem~\cite{liang2006problem,michalewicz2000testcase}. As a result, when an optimizer performs poorly on a fixed instance, it is often difficult to determine whether the failure was primarily caused by objective-side difficulty~\cite{hansen2009bbob,yazdani2023gnbg}, by the scarcity of feasible points~\cite{michalewicz2000testcase}, by restrictive active constraints~\cite{nocedal2006numerical}, by nonlinear feasible-region geometry~\cite{michalewicz2000testcase,liang2006problem}, or by numerical properties of the constraint functions~\cite{runarsson2000stochastic}. Fixed suites are therefore indispensable for reproducible comparison~\cite{hansen2021coco,hellwig2019benchmarking}, but less convenient when the research goal is to isolate and vary specific mechanisms of constrained search~\cite{michalewicz2000testcase,yazdani2023gnbg}.

Generated and configurable benchmarks offer a complementary approach to fixed test suites~\cite{michalewicz2000testcase,yazdani2023gnbg}. Instead of relying only on a catalogue of predefined functions, a generator can create families of problem instances whose properties are deliberately varied~\cite{michalewicz2000testcase,gandomi2023gnbgtestsuite}. In continuous numerical optimization, configurable objective generators have shown how a unified parametric construction can support controlled studies of landscape properties such as modality, separability, variable interaction, dimensionality, conditioning, and basin structure~\cite{yazdani2023gnbg,gandomi2023gnbgtestsuite,salgotra2025gnbgii, salgotra2026gnbgiii}. In constrained optimization, earlier test-case generators also recognized the importance of controlling feasibility-related properties, including feasible-region size, constraint types, active constraints, and feasible-space topology~\cite{michalewicz2000testcase,schmidt2000tcg2}. These developments show that configurability is valuable on both the objective side and the constraint side~\cite{michalewicz2000testcase,yazdani2023gnbg}. The remaining need is to make the feasible region itself a modular and inspectable benchmark object: one that can be varied independently of the objective model, recombined across diagnostic settings, and checked against its intended properties.

This motivation leads to the central premise of this work: the feasible region should be treated as a configurable benchmark object, not merely as a fixed consequence of a predefined constrained test function~\cite{michalewicz2000testcase,schmidt2000tcg2}. We introduce GNBG-C, the Generalized Numerical Benchmark Generator for Constrained Continuous Optimization. GNBG-C generates inequality-constraint layers that can be coupled with a GNBG objective model to produce constrained problem instances. Rather than hard-coding constraints into a fixed test function, it makes feasibility-side structure configurable. This allows researchers to vary the restrictiveness of the feasible region, the structure of constraint boundaries, the pressure imposed by active constraints, the degree of variable involvement, the numerical scaling of constraint values, and the presence of redundant constraints~\cite{michalewicz2000testcase,schmidt2000tcg2,runarsson2000stochastic,boneh1993constraint}.

The design of GNBG-C is modular. The objective model and the feasibility layer are conceptually separated: one can keep the objective landscape fixed while varying the generated feasible region, or apply similar feasibility regimes across different objective landscapes. In the implementation studied in this paper, the objective side is provided by GNBG, while GNBG-C adds the configurable constraint layer~\cite{yazdani2023gnbg,gandomi2023gnbgtestsuite}. This separation is important because it enables experiments that are difficult to construct with fixed constrained suites~\cite{hellwig2019benchmarking,liang2006problem}. For example, one can compare loose and strict feasible regions under the same objective landscape, increase the number of active constraints without changing the objective model, or test whether an optimizer is sensitive to dense variable coupling or ill-scaled constraint values while preserving other aspects of the benchmark~\cite{michalewicz2000testcase,runarsson2000stochastic}. Figure~\ref{fig:intro_objective_feasible_regions} provides a compact visual summary of this objective--feasibility separation.

A key design choice in GNBG-C is to generate constraints around a known GNBG reference solution~\cite{michalewicz2000testcase,schmidt2000tcg2}. The present study considers two known-optimum modes: GNBG-C-I, in which the known GNBG optimum \(x^\star\) is strictly feasible, and GNBG-C-B, in which \(x^\star\) remains feasible and one or more constraints are active at that point. These modes allow feasibility pressure to be varied without losing a known target for performance assessment~\cite{hansen2021coco,gandomi2023gnbgtestsuite}. This choice is not intended to cover every possible constrained scenario. In particular, it does not focus on cases where the unconstrained optimum is made infeasible and the constrained optimum shifts to an unknown location. Instead, GNBG-C targets controlled diagnostic benchmarking under known-reference conditions, where the effect of feasibility-side properties can be studied without sacrificing interpretability of the target solution~\cite{michalewicz2000testcase,gandomi2023gnbgtestsuite}.

The same emphasis on controllability also motivates auditability~\cite{hansen2021coco,hellwig2019benchmarking}. A generated benchmark instance should not only state its intended properties; it should allow the user to check whether those properties were actually achieved~\cite{hansen2021coco,gandomi2023gnbgtestsuite}. GNBG-C therefore records structural diagnostics associated with the generated constraint layer, including whether the reference solution is feasible, whether the intended active-set structure is obtained, whether the target feasibility regime is reached, and whether redundant constraints behave as expected. These diagnostics make the generated instances more transparent and help distinguish benchmark design from arbitrary constraint generation.

The main contributions of this work are summarized as follows:
\begin{itemize}
    \item We introduce GNBG-C, a benchmark-generation framework that treats the feasible region as a configurable component of constrained continuous optimization problems, rather than as a fixed attachment to a predefined test function.

    \item We develop a modular inequality-constraint layer for generating controlled feasibility-side regimes, covering feasible-region restrictiveness, boundary structure, active-set pressure, variable involvement, numerical scaling, and redundancy.

    \item We adopt a known-reference construction and accompanying structural diagnostics, allowing generated instances to preserve an attainable feasible solution while verifying whether the intended feasibility-side properties have been achieved.

    \item We define representative constrained benchmark regimes and provide baseline numerical experiments that illustrate how different feasibility conditions affect feasibility discovery, constraint handling, and objective improvement.
\end{itemize}
\section{Benchmark Design}
\label{sec:benchmark-design}

\begin{figure}[h]
    \centering
    \includegraphics[width=0.6\linewidth]{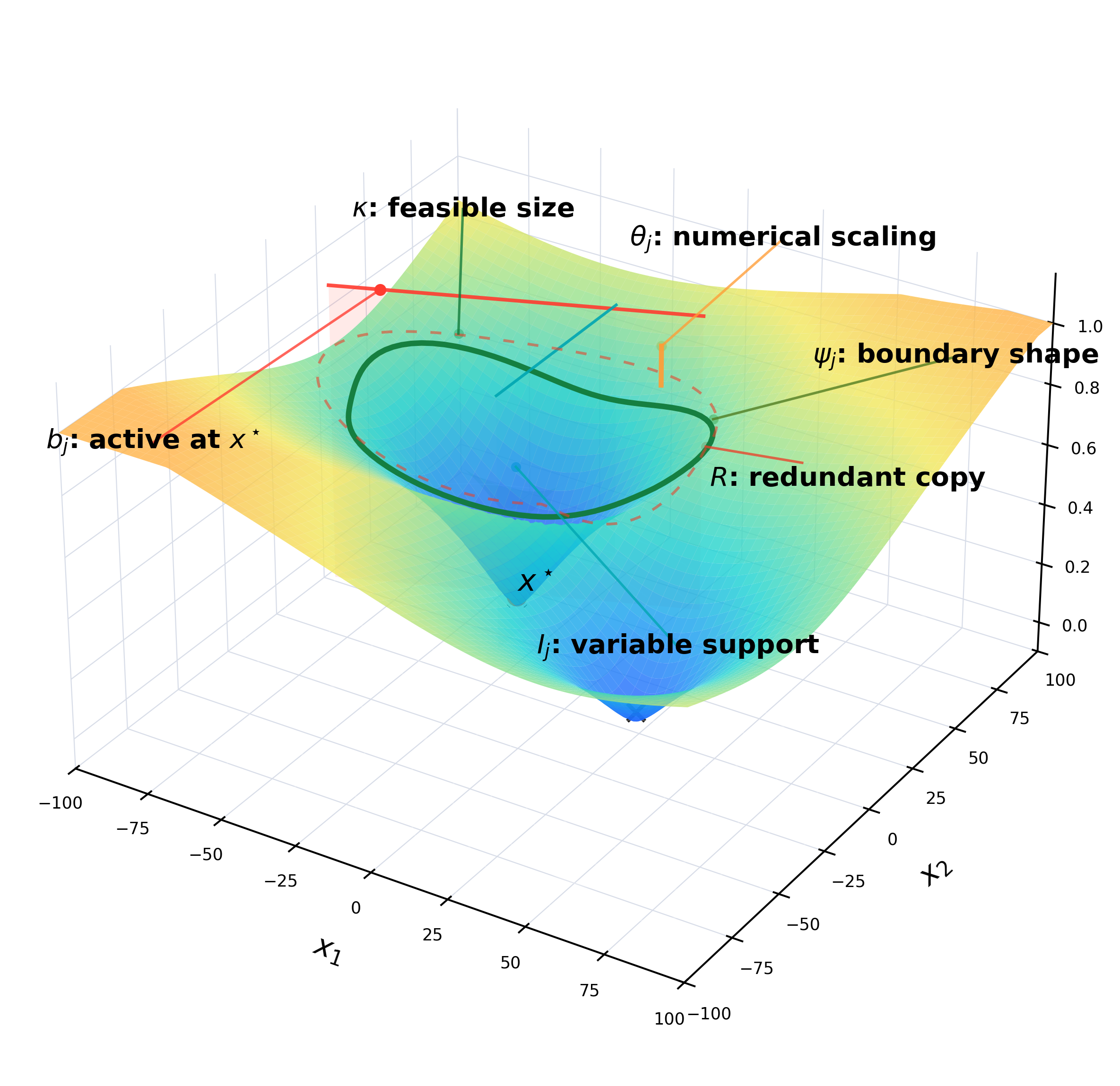}
    \caption{
    Conceptual view of GNBG-C as a configurable feasible-region layer attached to a GNBG objective landscape. 
    Starting from a fixed objective landscape and a known GNBG optimum \(x^\star\), the constraint layer acts as a controllable shell whose size, activity at the optimum, boundary geometry, variable participation, numerical scale, and redundancy can be adjusted through \(\kappa\), \(b_j\), \(\psi_j\), \(I_j\), \(\theta_j\), and \(\mathcal{R}\), respectively.
    }
    \label{fig:configurable-feasible-shell}
\end{figure}

\begin{figure*}[h]
    \centering
    \includegraphics[width=\linewidth]{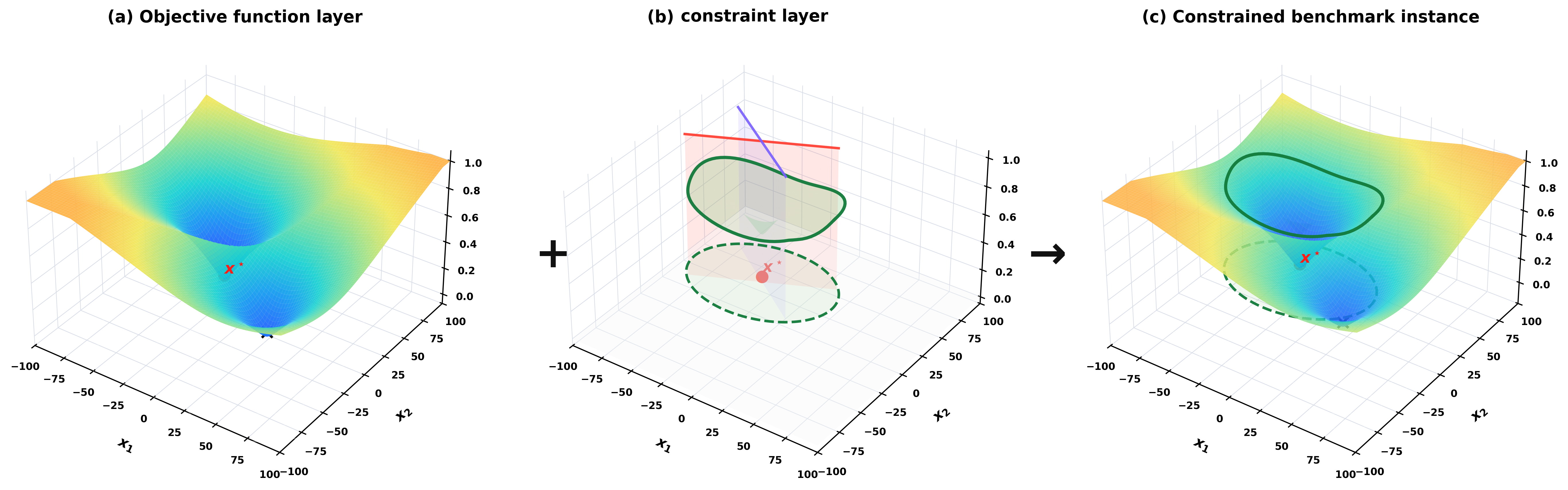}
    \caption{
    Layered construction of a GNBG-C benchmark instance. 
    The GNBG objective layer defines the base optimization landscape and the known reference optimum \(x^\star\)~\cite{yazdani2023gnbg,gandomi2023gnbgtestsuite}, while the GNBG-C constraint layer adds configurable feasibility-side structure around \(x^\star\). 
    Their composition yields a constrained GNBG instance in which objective-side properties and feasibility-side properties can be configured, diagnosed, and inspected separately.
    }
    \label{fig:layered-construction}
\end{figure*}

The feasible region now becomes the additional object of construction in
GNBG-C. Starting from a GNBG objective with a known global minimizer
\(x^\star\), GNBG-C adds an inequality-constraint layer that keeps this
point feasible while making the induced feasible region configurable, as
illustrated in Figure~\ref{fig:configurable-feasible-shell}. The layered
composition of the original objective and the proposed constraint
construction is summarized in Figure~\ref{fig:layered-construction}. The
construction supports both interior and boundary known-optimum
configurations by controlling the activity of the generated constraints
at \(x^\star\). A single centered constraint form organizes this layer,
with its parameters determining the feasible-region size, the activity of
constraints at the reference point, the shape of the boundaries, the
variables involved, the numerical scale of violations, and the amount of
redundant constraint-handling burden%
~\cite{hansen2021coco,hellwig2019benchmarking,liang2006problem,coello2002theoretical}.

To make this construction operational, let \(X=[l,u]^d\) be the box domain and let \(f:X\rightarrow\mathbb{R}\) be a GNBG objective function with known global minimizer \(x^\star\in X\). The generated constrained problem is

\begin{equation}
\label{eq:constrained-problem}
\min_{x\in X} f(x)
\quad \mathrm{s.t.} \quad
g_j(x)\leq 0,\; j=1,\ldots,m .
\end{equation}

These inequalities define
\(\Omega=\{x\in X\mid g_j(x)\leq 0,\; j=1,\ldots,m\}\)~\cite{liang2006problem,hellwig2019benchmarking}.
Since \(x^\star\) is a global minimizer of \(f\) on \(X\) and
\(x^\star\in\Omega\subseteq X\), it follows that
\(f(x^\star)\leq f(x)\) for every \(x\in\Omega\).
Hence, \(x^\star\) is also a global minimizer of the constrained problem.
This preserves the known-reference structure of the benchmark while allowing feasibility-side properties to vary.

The present framework supports two known-optimum modes. In \textbf{GNBG-C-I} (interior known optimum), the original GNBG optimum is strictly feasible with respect to every generated constraint,
\[
g_j(x^\star)<0,
\qquad j=1,\ldots,m.
\]
In \textbf{GNBG-C-B} (boundary known optimum), the original GNBG optimum remains feasible with respect to every generated constraint, while one or more constraints are active at that point,
\[
g_j(x^\star)\leq 0,
\qquad j=1,\ldots,m,
\qquad
\exists j\in\{1,\ldots,m\}: g_j(x^\star)=0.
\]

To describe a candidate solution \(x\), the benchmark reports four quantities: the objective value \(f(x)\), the constraint vector \(g(x)=(g_1(x),\ldots,g_m(x))\), the feasibility status, and the aggregate violation
\(\mathrm{CV}(x)=\sum_{j=1}^{m}\max\{g_j(x),0\}\)~\cite{deb2000efficient,hellwig2019benchmarking}.

\subsection{Optimum-preserving constraint-layer form}
\label{subsec:optimum-preserving-layer}

The preservation of \(x^\star\) is encoded directly in the constraint form. Each constraint is expressed relative to the reference point, so feasibility follows from the algebraic structure of the layer rather than from an additional feasibility-enforcement step. For constraint \(j\), the subset \(I_j\subseteq\{1,\ldots,d\}\) specifies the variables on which the boundary depends. The generated layer is written as

\begin{equation}
\label{eq:general-constraint-layer}
\begin{aligned}
g_j(x)
&=
\psi_j\!\left(z_{I_j}(x);\theta_j\right)-\kappa b_j,
\qquad j=1,\ldots,m,\\
z_i(x)
&=
\frac{x_i-x_i^\star}{\max\{u_i-l_i,\epsilon_s\}},
\qquad i=1,\ldots,d .
\end{aligned}
\end{equation}

Here, \(z_i(x)\) measures displacement from the GNBG optimum relative to the box scale, where \(\epsilon_s>0\) keeps the normalization well defined when a coordinate span degenerates. The function \(\psi_j\) is the centered family-specific component of the constraint, \(\theta_j\) contains its parameters, \(b_j\geq 0\) sets the slack at the reference point, and \(\kappa>0\) is a shared calibration multiplier. Since \(\psi_j(0;\theta_j)=0\), substituting \(x=x^\star\) gives \(g_j(x^\star)=-\kappa b_j\leq 0\). Thus, the known-optimum property and the controllability of the GNBG-C layer come from the same centered form. In particular, \(b_j>0\) leaves constraint \(j\) inactive at \(x^\star\), while \(b_j=0\) makes it active at \(x^\star\). Accordingly, GNBG-C-I is obtained by assigning \(b_j>0\) to every generated constraint. GNBG-C-B is obtained by assigning \(b_j=0\) to one or more constraints while maintaining \(b_j\geq 0\) for all remaining constraints.

Each element in \cref{eq:general-constraint-layer} corresponds to a distinct control over the generated feasible region. The multiplier \(\kappa\) calibrates feasible-region size through constraints with positive slack at \(x^\star\), but it does not affect zero-slack active constraints; the slack \(b_j\) determines whether constraint \(j\) is active at \(x^\star\) or leaves a margin there; \(\psi_j\) specifies the boundary family; \(I_j\) selects the participating variables; and \(\theta_j\) contains the family-specific shape and scale parameters. These parameter--property correspondences are
summarized in Table~\ref{tab:constraint-parameter-map}.

\begin{table}[t]
\centering
\caption{Parameter map of the centered GNBG-C constraint layer and the benchmark property controlled by each element.}
\label{tab:constraint-parameter-map}
\small
\begin{tabular}{p{0.21\columnwidth}p{0.69\columnwidth}}
\toprule
Element & Controlled benchmark property \\
\midrule
\(\kappa\) &
Shared calibration of feasible-region size through positive-slack constraints; zero-slack active constraints are unaffected. \\[0.2em]

\(b_j\) &
Local activity and margin of constraint \(j\) at \(x^\star\). \\[0.2em]

\(\psi_j\) &
Boundary family and geometric complexity. \\[0.2em]

\(I_j\) &
Sparse or dense variable participation in constraint \(j\). \\[0.2em]

\(\theta_j\) &
Family parameters, coefficient magnitudes, and numerical scaling. \\[0.2em]

\(\mathcal{R}\) &
Redundant constraints that increase constraint-handling burden without tightening the feasible set. \\
\bottomrule
\end{tabular}
\end{table}

\subsection{Controls exposed by the centered layer}
\label{subsec:controls-exposed-centered-layer}

The controls can now be read directly from the centered form. The shared multiplier \(\kappa\) provides a calibration control for the feasible-region ratio through the positive-slack terms: by shifting the centered inequalities with \(b_j>0\), it changes how much of the search domain satisfies the generated constraint set. Zero-slack active constraints are not shifted by \(\kappa\), so in active-boundary configurations the attainable feasible-region ratio may also be limited by those constraints. The resulting feasible-region size is described through the feasible-region ratio \(\rho=\mathrm{vol}(\Omega)/\mathrm{vol}(X)\)~\cite{hellwig2019benchmarking}. For mixed nonlinear families, \(\rho\) is generally not available in closed form, so the calibration stage estimates it from samples \(\{x^{(n)}\}_{n=1}^{N}\)~\cite{hellwig2019benchmarking}:

\begin{equation}
\label{eq:estimated-feasibility-ratio}
\hat{\rho}(\kappa)=
\frac{1}{N}\sum_{n=1}^{N}
\mathbb{I}\!\left[
g_j(x^{(n)};\kappa)\leq \epsilon_{\mathrm{feas}},
\; \forall j\in\{1,\ldots,m\}
\right].
\end{equation}
The shared multiplier \(\kappa\) is selected so that
\(\hat{\rho}(\kappa)\) approaches the requested target
\(\rho_{\mathrm{target}}\) within an absolute tolerance \(\tau\).
Because \(\kappa\) shifts only positive-slack constraints, the requested
ratio may not always be attainable in boundary-active configurations.
When the target cannot be reached, the evaluated multiplier producing the
closest achieved ratio is retained together with a calibration status.
The detailed procedure and the experimental settings are reported in
\cref{alg:feasibility-calibration} and the accompanying text.

A global feasible fraction alone does not describe the local structure near the optimum. Two instances can have similar values of \(\hat{\rho}\) while presenting different local restrictions around \(x^\star\). This local structure is controlled by the slack terms. From \cref{eq:general-constraint-layer}, \(b_j=0\) implies \(g_j(x^\star)=0\), whereas \(b_j>0\) implies \(g_j(x^\star)<0\). Assigning zero slack to selected constraints therefore determines which boundaries pass through the reference optimum, while positive slack leaves the remaining constraints inactive at that point. The active set is \(\mathcal{A}(x^\star)=\{j\in\{1,\ldots,m\}\mid b_j=0\}\), and its size can be controlled without moving \(x^\star\) outside the feasible region. This controls whether the reference optimum is strictly feasible or boundary-active; the local geometric effect of those active constraints still depends on the selected family and parameters.

To avoid degenerate active boundaries, GNBG-C screens candidate
constraints using their local gradient in the normalized coordinate
system. For each smooth candidate constraint, define
\[
\nu_j
=
\left\|
\nabla_z\psi_j(0;\theta_j)
\right\|_2 .
\]
A candidate is eligible for the active set only when
\(\nu_j>\epsilon_{\nabla}\), where
\(\epsilon_{\nabla}=10^{-8}\) in the present experiments. Because the
mapping between \(x\) and \(z\) is an affine coordinate-wise scaling,
this criterion excludes an active boundary whose local gradient
vanishes at \(x^\star\). If fewer eligible candidates than the requested active-set size are generated, the candidate instance is rejected before inclusion in the
benchmark set. The active-gradient screening outcome is
also retained in the instance diagnostics.

Once the position of a boundary relative to \(x^\star\) is fixed, its shape becomes the next source of feasible-region structure. This is the role of \(\psi_j\). Linear constraints generate half-space boundaries, convex quadratic constraints produce regular curved boundaries, nonconvex quadratic constraints introduce indefinite curvature, and higher-order or oscillatory forms create more irregular violation profiles. The implemented centered families used in GNBG-C are summarized in \cref{tab:constraint-families}.

\begin{table}[h]
\centering
\caption{Implemented GNBG-C constraint families through the centered function \(\psi_j\).}
\label{tab:constraint-families}
\small
\begin{tabular}{p{0.29\columnwidth}p{0.61\columnwidth}}
\toprule
Family & Centered raw form \(\psi_j(z;\theta_j)\) \\
\midrule
Linear &
\(a_j^\top z\) \\[0.2em]

Convex quadratic &
\(a_j^\top z+z^\top Q_jz,\; Q_j\succeq 0\) \\[0.2em]

Nonconvex quadratic &
\(a_j^\top z+z^\top Q_jz,\; Q_j\) indefinite \\[0.2em]

Polynomial &
\(\sum_{r=1}^{R_j} c_{jr}(u_{jr}^{\top}z)^{p_{jr}}\) \\[0.2em]

Transcendental &
\(\sum_{r=1}^{R_j} c_{jr}
[\sin(\omega_{jr}u_{jr}^{\top}z+\phi_{jr})-\sin(\phi_{jr})]\) \\
\bottomrule
\end{tabular}
\end{table}

The algebraic type of a boundary describes its shape, but not the portion of the decision vector on which that shape acts. The subset \(I_j\) specifies which variables are involved in the \(j\)-th boundary. A sparse constraint localizes feasibility changes to fewer coordinates, whereas a dense constraint couples many coordinates and makes feasible movement more global. If an involvement rate \(r_{\mathrm{inv}}\in(0,1]\) is specified, the support size can be written as \(|I_j|=\max\{1,\mathrm{round}(r_{\mathrm{inv}}d)\}\). Across all constraints, the collection \(\mathcal{I}=\{I_j\}_{j=1}^{m}\) determines the aggregate variable-participation pattern of the feasible region.

Even with the family and participating variables fixed, the same constraint can produce violation values on very different numerical scales. This remaining degree of control is carried by \(\theta_j\). It includes coefficient magnitudes, spectra of quadratic forms, polynomial coefficients, and frequency-related parameters in oscillatory constraints. To express this control without tying the benchmark definition to fixed numerical constants, one may write \(\theta_j=(\bar{\theta}_j,\gamma_j)\), where \(\bar{\theta}_j\) contains unscaled family parameters and \(\gamma_j>0\) controls magnitude. A balanced regime samples \(\gamma_j\) from a moderate bounded range, whereas an ill-scaled regime samples it on a logarithmic scale, for example \(\gamma_j=\beta_s^{\xi_j}\), with \(\xi_j\sim\mathcal{U}(\eta_{\min},\eta_{\max})\) and \(\beta_s>1\). This changes the numerical scale of violations without requiring a different boundary family, which is important for solvers that rank, penalize, or repair candidates using violation magnitudes~\cite{deb2000efficient,hellwig2019benchmarking}.

The last control changes the constraint set without reshaping the feasible
region itself. An implied inequality can be generated from an existing
parent \(g_p(x)\leq 0\) as
\[
g_r(x)=g_p(x)-\delta_r,
\qquad \delta_r>0.
\]
Since \(g_p(x)\leq 0\Rightarrow g_r(x)\leq 0\), the child constraint
cannot tighten the feasible set. Hence, adding such children leaves
\(\Omega\) unchanged. Nevertheless, the aggregate violation presented
to an optimizer may change for infeasible points. In particular, adding
one redundant child changes the reported violation by
\[
\mathrm{CV}_{\mathrm{with}\ r}(x)
=
\mathrm{CV}_{\mathrm{without}\ r}(x)
+
\max\{g_p(x)-\delta_r,0\}.
\]
Thus, whenever \(g_p(x)>\delta_r\), the redundant child increases the
aggregate violation even though it introduces no new feasible boundary.
This is an intended solver-facing effect: the redundant set
\(\mathcal{R}\) tests the sensitivity of penalty-, ranking-, and
violation-based constraint-handling mechanisms to implied feasibility
information~\cite{coello2002theoretical,hellwig2019benchmarking}.

\subsection{Generation, calibration, and diagnostics}
\label{subsec:generation-calibration-diagnostics}

For a generated GNBG-C instance, specifying the intended configuration is not sufficient; the realized instance must also be auditable~\cite{hansen2021coco,hellwig2019benchmarking}. This is especially important because the feasible-region ratio is estimated from samples, and because sampled families, variable supports, parameter scales, and redundant children can introduce instance-level variation. The generation procedure therefore records both the configuration used to construct the constraint layer and the diagnostics obtained after construction.

\begin{algorithm}[t]
\caption{GNBG-C constraint-layer generation}
\label{alg:constraint-layer-generation}
\begin{algorithmic}[1]

\Require Box domain \(X=[l,u]^d\); reference optimum \(x^\star\);
constraint configuration \(\mathcal{C}\); active-gradient threshold
\(\epsilon_{\nabla}\)

\Ensure Accepted constraint set \(\mathcal{G}\) and diagnostics
\(\mathcal{D}\), or a rejection status

\State Read the number of primary constraints, family mix, requested
active-set size \(a_{\mathrm{req}}\), and redundancy level from
\(\mathcal{C}\).

\State Initialize the primary and redundant sets as
\(\mathcal{P}\leftarrow\emptyset\) and
\(\mathcal{R}\leftarrow\emptyset\).

\For{each primary constraint \(j\)}
    \State Sample a family \(\psi_j\), variable subset \(I_j\),
    parameters \(\theta_j\), and positive base slack \(b_j\).
    \State Compute the normalized local-gradient measure
\(\nu_j\leftarrow\|\nabla_z\psi_j(0;\theta_j)\|_2\).
\EndFor

\State Form the eligible set
\(\mathcal{E}\leftarrow
\{j:\nu_j>\epsilon_{\nabla}\}\).

\If{\(|\mathcal{E}|<a_{\mathrm{req}}\)}
    \State Record the number of missing eligible active constraints.
    \State \Return a degenerate-active-set rejection status.
\EndIf

\State Select an active set
\(\mathcal{A}\subseteq\mathcal{E}\) with
\(|\mathcal{A}|=a_{\mathrm{req}}\).

\For{each primary constraint \(j\)}
    \If{\(j\in\mathcal{A}\)}
        \State Set \(b_j\leftarrow0\).
    \EndIf
\EndFor

\State Calibrate the shared multiplier \(\kappa\) using
\cref{alg:feasibility-calibration}.

\For{each primary constraint \(j\)}
    \State Construct
    \(g_j(x)=\psi_j(z_{I_j}(x);\theta_j)-\kappa b_j\)
    and add it to \(\mathcal{P}\).
\EndFor

\State Append redundant children
\(g_r(x)=g_p(x)-\delta_r\), with \(\delta_r>0\), as requested by
\(\mathcal{C}\).

\State Form the final set
\(\mathcal{G}\leftarrow\mathcal{P}\cup\mathcal{R}\).

\State Validate reference feasibility, active-set realization,
active-gradient nondegeneracy, achieved feasibility ratio, family
counts, scaling regime, and redundancy implication.

\State Record the diagnostics \(\mathcal{D}\).

\State \Return \(\mathcal{G}\), \(\mathcal{D}\), and an accepted status.

\end{algorithmic}
\end{algorithm}
The calibration step in \cref{alg:constraint-layer-generation} is detailed in \cref{alg:feasibility-calibration}.

\begin{algorithm}[t]
\caption{Feasibility-ratio calibration}
\label{alg:feasibility-calibration}
\begin{algorithmic}[1]
\Require Fixed calibration sample \(\mathcal{S}\); primary constraint
templates \(\{\psi_j,\theta_j,I_j,b_j\}_{j=1}^{m}\); target ratio
\(\rho_{\mathrm{target}}\); tolerance \(\tau\); initial interval
\([\kappa_{0}^{-},\kappa_{0}^{+}]\); hard bounds
\([\kappa_{\min},\kappa_{\max}]\); expansion factor \(\gamma\); maximum
iterations \(T_{\max}\)
\Ensure Multiplier \(\kappa\), achieved ratio \(\hat{\rho}\), and calibration status
\State Set \(\kappa_{\mathrm{lo}}\leftarrow\kappa_{0}^{-}\) and
\(\kappa_{\mathrm{hi}}\leftarrow\kappa_{0}^{+}\).
\State Evaluate both endpoints on \(\mathcal{S}\) and retain the candidate
closest to \(\rho_{\mathrm{target}}\).
\While{\(\hat{\rho}(\kappa_{\mathrm{hi}})<\rho_{\mathrm{target}}
\ \land\ \kappa_{\mathrm{hi}}<\kappa_{\max}\)}
    \State Set \(\kappa_{\mathrm{hi}}\leftarrow
    \min\{\gamma\kappa_{\mathrm{hi}},\kappa_{\max}\}\).
    \State Evaluate the new endpoint and update the closest candidate.
\EndWhile
\While{\(\hat{\rho}(\kappa_{\mathrm{lo}})>\rho_{\mathrm{target}}
\ \land\ \kappa_{\mathrm{lo}}>\kappa_{\min}\)}
    \State Set \(\kappa_{\mathrm{lo}}\leftarrow
    \max\{\kappa_{\mathrm{lo}}/\gamma,\kappa_{\min}\}\).
    \State Evaluate the new endpoint and update the closest candidate.
\EndWhile
\If{\(\rho_{\mathrm{target}}\) is not bracketed}
    \State \Return the closest candidate and an out-of-range status.
\EndIf
\For{\(t=1,\ldots,T_{\max}\)}
    \State Set \(\kappa\leftarrow
    \sqrt{\kappa_{\mathrm{lo}}\kappa_{\mathrm{hi}}}\).
    \State Evaluate \(\hat{\rho}(\kappa)\) on \(\mathcal{S}\) and update the
    closest candidate.
    \If{\(|\hat{\rho}(\kappa)-\rho_{\mathrm{target}}|\leq\tau\)}
        \State \Return \(\kappa\), \(\hat{\rho}(\kappa)\), and success.
    \ElsIf{\(\hat{\rho}(\kappa)<\rho_{\mathrm{target}}\)}
        \State Set \(\kappa_{\mathrm{lo}}\leftarrow\kappa\).
    \Else
        \State Set \(\kappa_{\mathrm{hi}}\leftarrow\kappa\).
    \EndIf
\EndFor
\State \Return the closest candidate and a maximum-iterations status.
\end{algorithmic}
\end{algorithm}

\Cref{alg:feasibility-calibration} operates on a fixed calibration sample
\(\mathcal{S}=\{x^{(n)}\}_{n=1}^{N_{\mathrm{cal}}}\). In the experiments,
\(N_{\mathrm{cal}}=2048\) scrambled Sobol points were sampled over the box
domain \(X\). For an instance generated with constraint seed \(s_c\), the
Sobol sample used seed \(s_c+101\). The same sample was reused for every
evaluated value of \(\kappa\), thereby preventing resampling variation from
affecting the interval search. A sampled point was classified as feasible
when
\[
g_j(x)\leq\epsilon_{\mathrm{feas}}=10^{-10},
\qquad j=1,\ldots,m.
\]

The initial calibration interval was
\([\kappa_{0}^{-},\kappa_{0}^{+}]=[0.25,4]\), the hard bounds were
\([\kappa_{\min},\kappa_{\max}]=[10^{-8},10^{6}]\), the expansion factor was
\(\gamma=2\), and the maximum number of multiplicative-bisection iterations
was \(T_{\max}=20\). All interval midpoints were computed geometrically as
\(\sqrt{\kappa_{\mathrm{lo}}\kappa_{\mathrm{hi}}}\). For the \(d=10\)
regimes, the absolute calibration tolerances \(\tau\) for loose, medium,
strict, active-high, mostly nonlinear, dense, ill-scaled, and
high-redundancy configurations were \(0.050\), \(0.025\), \(0.004\),
\(0.002\), \(0.010\), \(0.002\), \(0.015\), and \(0.025\), respectively.

Increasing \(\kappa\) weakly relaxes every positive-slack constraint, so the
estimated feasibility ratio is nondecreasing with respect to \(\kappa\).
Zero-slack active constraints, however, are unaffected by \(\kappa\) and may
therefore limit the attainable ratio in GNBG-C-B instances. If the target
ratio was outside the estimated attainable range, if the adaptive interval
reached either hard bound, or if \(T_{\max}\) iterations were completed
without satisfying \(\tau\), the evaluated candidate minimizing $|\hat{\rho}(\kappa)-\rho_{\mathrm{target}}|$ was retained. Ties were resolved in favor of the candidate whose multiplier
was closest to \(\kappa=1\).

The recorded diagnostics include whether \(x^\star\) is feasible, the
total violation at \(x^\star\), the requested and realized active-set
sizes, the number of weak active constraints, the realized family
counts, the target and achieved feasibility ratios, the selected value
of \(\kappa\), the absolute calibration error, the calibration status,
and the validity of redundant implications. These quantities are
benchmark metadata rather than solver outputs and allow empirical
results to be interpreted through the realized properties of the
constructed feasible region
~\cite{hansen2021coco,hellwig2019benchmarking,gandomi2023gnbgtestsuite}.

\subsection{Objective layer used in this work}
\label{subsec:objective-layer-used}

With the feasible-region layer specified and diagnosed independently, the remaining component is the GNBG objective layer to which it is attached. In this work, GNBG is the base generator being extended: it provides configurable continuous landscapes with controllable objective-side properties and a known reference optimum~\cite{yazdani2023gnbg}. The constraint layer then adds feasibility-side controls around that optimum while leaving the GNBG objective definition unchanged.

GNBG defines its objective as the lower envelope of parameterized components~\cite{yazdani2023gnbg}. For component (k), let \(m_k\) be the center, \(\sigma_k\) the component minimum value, \(H_k\) a diagonal scaling matrix, \(R_k\) a rotation matrix, \(\lambda_k\) a basin-growth parameter, and \(T_k\) the element-wise nonlinear transformation used in GNBG. A compact form is
\begin{equation}
\label{eq:gnbg-objective}
\begin{aligned}
f(x)
&=
\min_{k\in\{1,\ldots,o\}}
\left[
\sigma_k+
\left(
y_k(x)^\top H_k y_k(x)
\right)^{\lambda_k}
\right],\\
y_k(x)&=T_k(R_k(x-m_k)),\qquad x\in X .
\end{aligned}
\end{equation}

The GNBG parameters control objective-side properties such as modality, basin morphology, conditioning, variable interaction, asymmetry, and component structure~\cite{yazdani2023gnbg,gandomi2023gnbgtestsuite}. These parameters remain separate from the proposed constraint layer, allowing objective-side and feasibility-side properties to be varied without conflating their roles in the generated constrained instance.

\section{Instance Diagnostics and Baseline Evaluation}
\label{sec:diagnostic-evaluation}

A configurable benchmark should not only generate diverse problem
instances; it should also make the source of each difficulty visible and
interpretable~\cite{hansen2021coco,hellwig2019benchmarking}. This section
examines GNBG-C from two complementary perspectives. First, controlled
visual comparisons are used to show how individual elements of the
constraint layer affect the feasible region and the information supplied
to a constraint-handling method. The generated regimes are then evaluated
with representative constrained optimizers to determine whether these
structural differences lead to distinct patterns of feasibility discovery
and best-feasible objective improvement. The purpose is diagnostic rather
than competitive: the experiments are intended to connect optimizer
behaviour to identifiable properties of the generated constraints, not to
establish a universal ranking of algorithms
~\cite{hellwig2019benchmarking,demsar2006statistical}.

\subsection{Generated regimes and structural diagnostics}
\label{subsec:generated-regimes}

The comparisons in this subsection focus on the observable effect of the
GNBG-C settings. For each comparison, the property of interest is varied
while the objective landscape, the known optimum, and unrelated parts of
the constraint construction are kept fixed whenever possible. This makes
it possible to attribute the visible change to the selected feasibility-side
property rather than to a simultaneous change in the underlying objective.

\begin{figure}[h]
\centering

\captionsetup[subfigure]{
    font={bf,scriptsize},
    labelfont=bf,
    justification=centering,
    singlelinecheck=false,
    skip=2pt
}

\setlength{\tabcolsep}{1.2pt}
\renewcommand{\arraystretch}{0.96}

\begin{tabular}{@{}ccc@{}}

\begin{subfigure}[t]{0.315\linewidth}
    \centering
    \includegraphics[width=\linewidth]{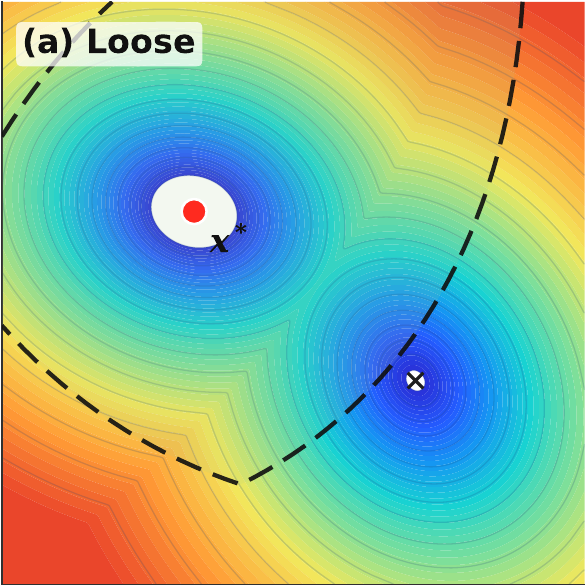}
    \caption{\protect\boldmath
    $\rho_{\mathrm{target}}=0.58$\\[3pt]
    $\hat{\rho}=0.56,\ \kappa=7.55$}
\end{subfigure}
&
\begin{subfigure}[t]{0.315\linewidth}
    \centering
    \includegraphics[width=\linewidth]{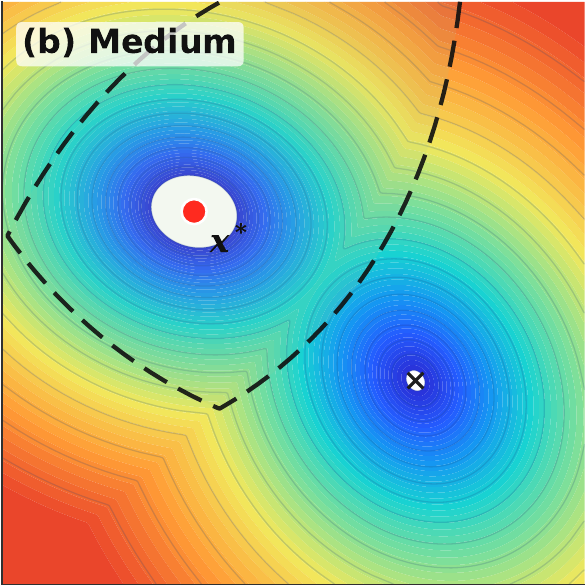}
    \caption{\protect\boldmath
    $\rho_{\mathrm{target}}=0.32$\\[3pt]
    $\hat{\rho}=0.35,\ \kappa=5.06$}
\end{subfigure}
&
\begin{subfigure}[t]{0.315\linewidth}
    \centering
    \includegraphics[width=\linewidth]{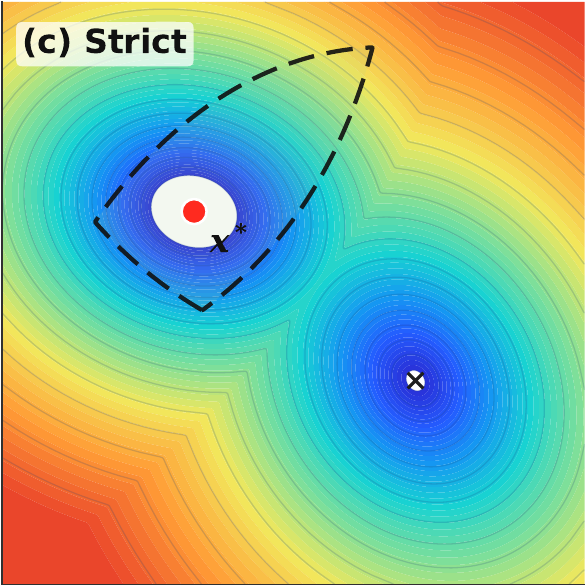}
    \caption{\protect\boldmath
    $\rho_{\mathrm{target}}=0.12$\\[3pt]
    $\hat{\rho}=0.11,\ \kappa=2.28$}
\end{subfigure}
\\[7pt]

\begin{subfigure}[t]{0.315\linewidth}
    \centering
    \includegraphics[width=\linewidth]{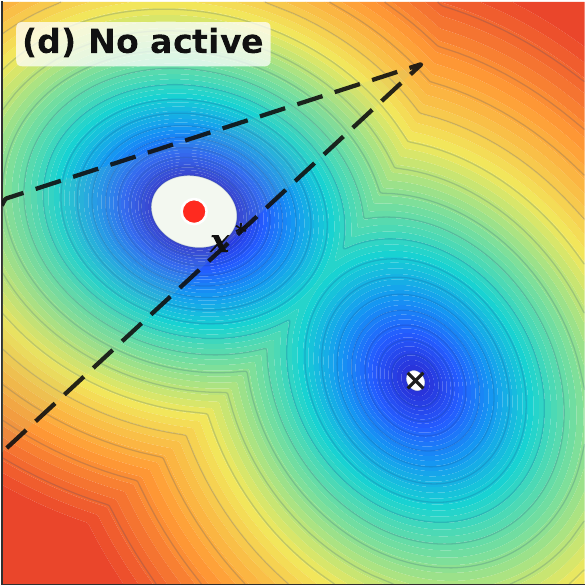}
    \caption{\protect\boldmath
    $|\mathcal{A}(x^\star)|=0$\\[3pt]
    $\hat{\rho}=0.16,\ \kappa=1.65$}
\end{subfigure}
&
\begin{subfigure}[t]{0.315\linewidth}
    \centering
    \includegraphics[width=\linewidth]{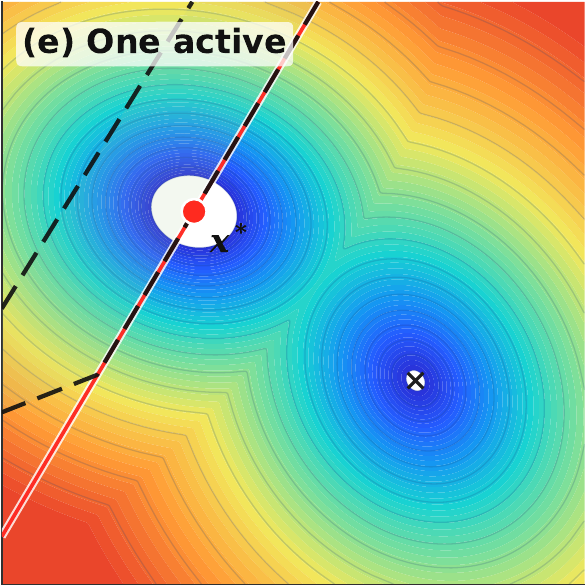}
    \caption{\protect\boldmath
    $|\mathcal{A}(x^\star)|=1$\\[3pt]
    $\hat{\rho}=0.15,\ \kappa=2.35$}
\end{subfigure}
&
\begin{subfigure}[t]{0.315\linewidth}
    \centering
    \includegraphics[width=\linewidth]{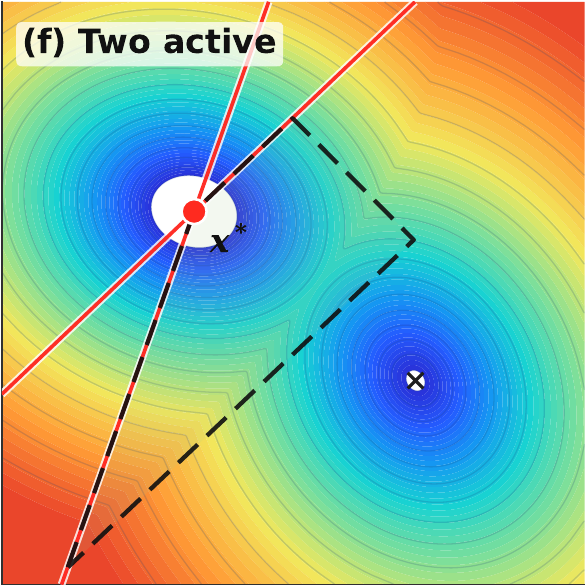}
    \caption{\protect\boldmath
    $|\mathcal{A}(x^\star)|=2$\\[3pt]
    $\hat{\rho}=0.15,\ \kappa=2.35$}
\end{subfigure}
\\[7pt]

\begin{subfigure}[t]{0.315\linewidth}
    \centering
    \includegraphics[width=\linewidth]{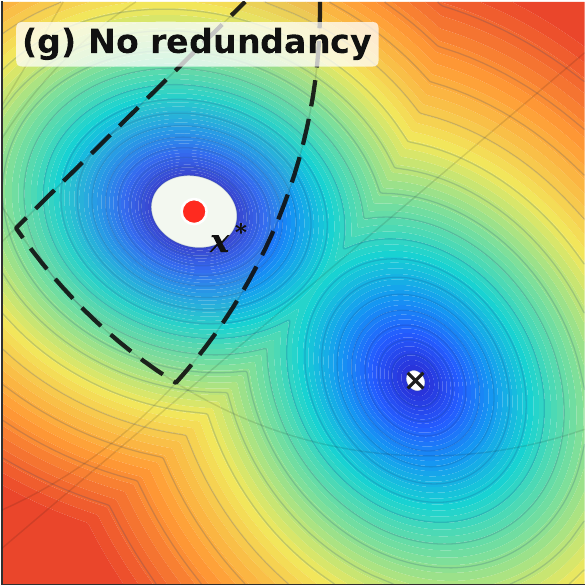}
    \caption{\protect\boldmath
    $m_0=6,\ |\mathcal{R}|=0$\\[3pt]
    $\hat{\rho}=0.18$}
\end{subfigure}
&
\begin{subfigure}[t]{0.315\linewidth}
    \centering
    \includegraphics[width=\linewidth]{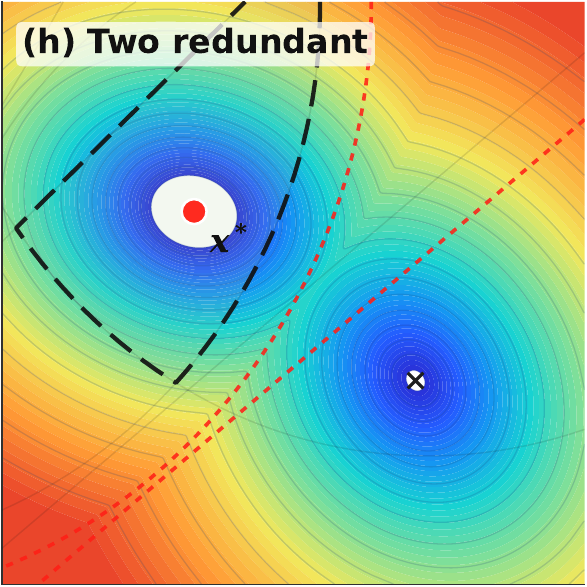}
    \caption{\protect\boldmath
    $m_0=6,\ |\mathcal{R}|=2$\\[3pt]
    $\hat{\rho}=0.18,\ \mathrm{changed}=0$}
\end{subfigure}
&
\begin{subfigure}[t]{0.315\linewidth}
    \centering
    \includegraphics[width=\linewidth]{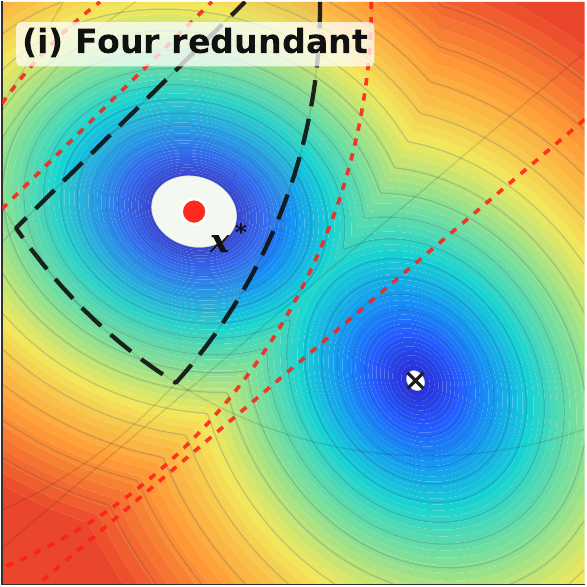}
    \caption{\protect\boldmath
    $m_0=6,\ |\mathcal{R}|=4$\\[3pt]
    $\hat{\rho}=0.18,\ \mathrm{changed}=0$}
\end{subfigure}

\end{tabular}

\vspace{-3pt}
\caption{Two-dimensional views of feasibility-ratio calibration,
active-set pressure, and redundancy. The first row varies the target
feasible fraction, the second varies the number of constraints active at
\(x^\star\) while keeping the achieved fraction comparable, and the
third adds implied constraints without changing sampled feasibility;
\(\mathrm{changed}=0\) indicates that no sampled classification changed.}
\label{fig:merged-config-active-redundancy}
\vspace{-8pt}
\end{figure}

The most immediate effect is the amount of admissible search space. In the
first row of Figure~\ref{fig:merged-config-active-redundancy}, the requested
feasibility ratio is reduced from a loose to a strict setting. The achieved
ratio decreases accordingly, and the visible feasible region contracts across
the three panels. This directly illustrates how feasibility-ratio calibration
controls the portion of the domain available to the optimizer
~\cite{jiao2019feasible,hellwig2019benchmarking}.

\begin{figure*}[t]
\centering

\captionsetup[subfigure]{
    font={bf,scriptsize},
    labelfont=bf,
    justification=centering,
    singlelinecheck=false,
    skip=2pt
}

\setlength{\tabcolsep}{1.2pt}
\renewcommand{\arraystretch}{0.96}

\begin{tabular}{@{}ccccc@{}}

\begin{subfigure}[t]{0.190\textwidth}
    \centering
    \includegraphics[width=\linewidth]{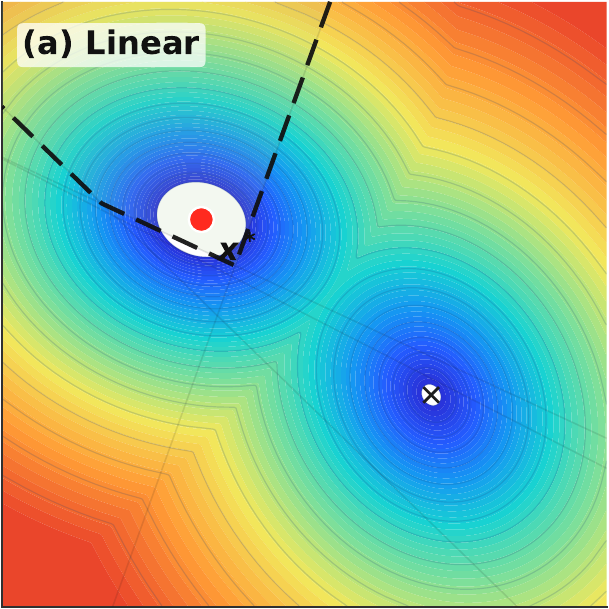}
    \caption{\protect\boldmath
    $\rho_{\mathrm{target}}=0.16$\\[2pt]
    $\hat{\rho}=0.16,\ \kappa=0.57$}
\end{subfigure}
&
\begin{subfigure}[t]{0.190\textwidth}
    \centering
    \includegraphics[width=\linewidth]{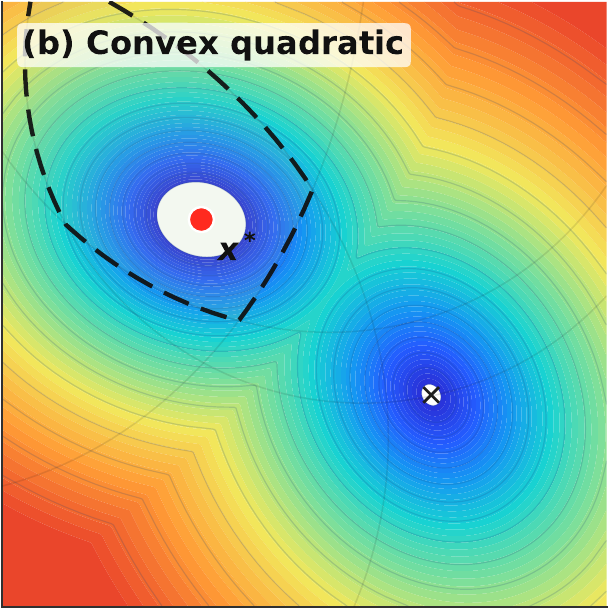}
    \caption{\protect\boldmath
    $\rho_{\mathrm{target}}=0.16$\\[2pt]
    $\hat{\rho}=0.16,\ \kappa=1.2$}
\end{subfigure}
&
\begin{subfigure}[t]{0.190\textwidth}
    \centering
    \includegraphics[width=\linewidth]{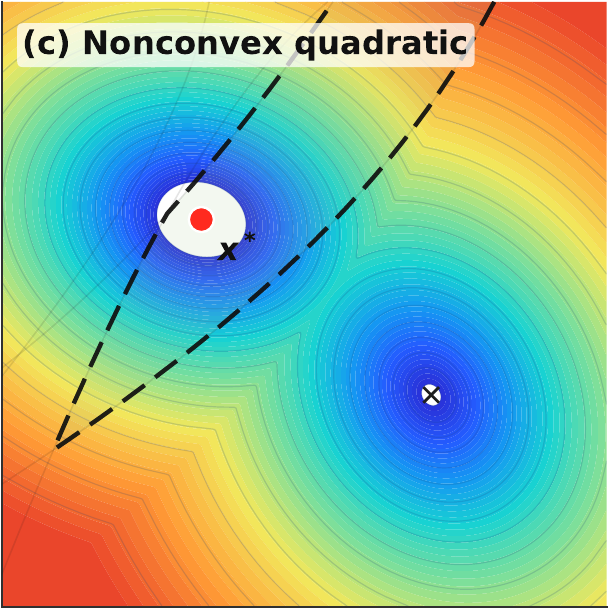}
    \caption{\protect\boldmath
    $\rho_{\mathrm{target}}=0.16$\\[2pt]
    $\hat{\rho}=0.16,\ \kappa=1.2$}
\end{subfigure}
&
\begin{subfigure}[t]{0.190\textwidth}
    \centering
    \includegraphics[width=\linewidth]{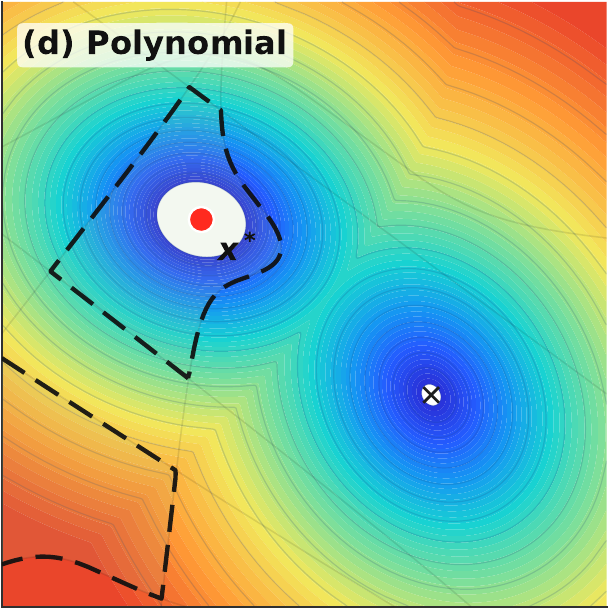}
    \caption{\protect\boldmath
    $\rho_{\mathrm{target}}=0.16$\\[2pt]
    $\hat{\rho}=0.16,\ \kappa=0.0083$}
\end{subfigure}
&
\begin{subfigure}[t]{0.190\textwidth}
    \centering
    \includegraphics[width=\linewidth]{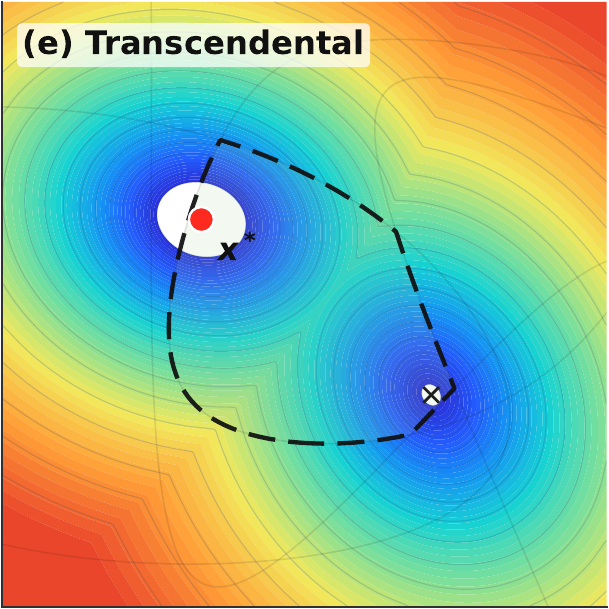}
    \caption{\protect\boldmath
    $\rho_{\mathrm{target}}=0.16$\\[2pt]
    $\hat{\rho}=0.17,\ \kappa=1.7$}
\end{subfigure}

\end{tabular}

\vspace{-4pt}
\caption{Effect of constraint-family choice on feasible-region geometry.}
\label{fig:boundary-family}
\vspace{-8pt}
\end{figure*}

Feasible-region size, however, does not fully describe the geometry near the
known optimum. The second row changes the number of constraints active at
\(x^\star\) from zero to two. With no active boundary, the reference point
lies in the interior of the feasible region; as active boundaries are
introduced, it becomes part of an increasingly restrictive local
intersection. The row therefore visualizes direct control over local boundary
pressure at the known optimum
~\cite{nocedal2006numerical,boyd2004convex}.

The third row increases the number of redundant constraints from zero to
four. The value \(\mathrm{changed}=0\) confirms that no sampled feasibility
classification changes as the redundant set grows, although the optimizer
receives additional inequalities and violation terms to aggregate, rank, or
penalize. The row therefore isolates the burden created by redundant
feasibility information
~\cite{boyd2004convex,coello2002theoretical,hellwig2019benchmarking}.

The next comparison varies the boundary family itself.
Figure~\ref{fig:boundary-family} moves from linear and convex-quadratic forms
to nonconvex-quadratic, polynomial, and transcendental constructions. The
resulting panels show how the same constraint layer can produce straight,
smoothly curved, indefinite, higher-order, and oscillatory feasibility
interfaces. These families are not intended to define an easy-to-hard
ordering; they expose different curvature and violation profiles to the
optimizer
~\cite{boyd2004convex,nocedal2006numerical,runarsson2000stochastic}.

\begin{figure}[h]
\centering

\captionsetup[subfigure]{
    font={bf,scriptsize},
    labelfont=bf,
    justification=centering,
    singlelinecheck=false,
    skip=2pt
}

\setlength{\tabcolsep}{1.2pt}
\renewcommand{\arraystretch}{0.94}

\begin{tabular}{@{}ccc@{}}

\begin{subfigure}[t]{0.315\linewidth}
    \centering
    \includegraphics[width=\linewidth]{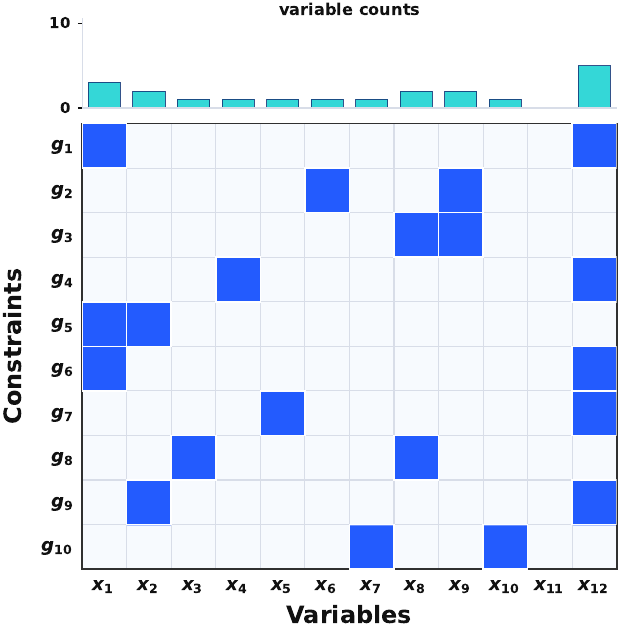}
    \caption{\protect\boldmath
    Sparse: $r_{\mathrm{inv}}=0.20$\\[2pt]
    density $=0.17$}
\end{subfigure}
&
\begin{subfigure}[t]{0.315\linewidth}
    \centering
    \includegraphics[width=\linewidth]{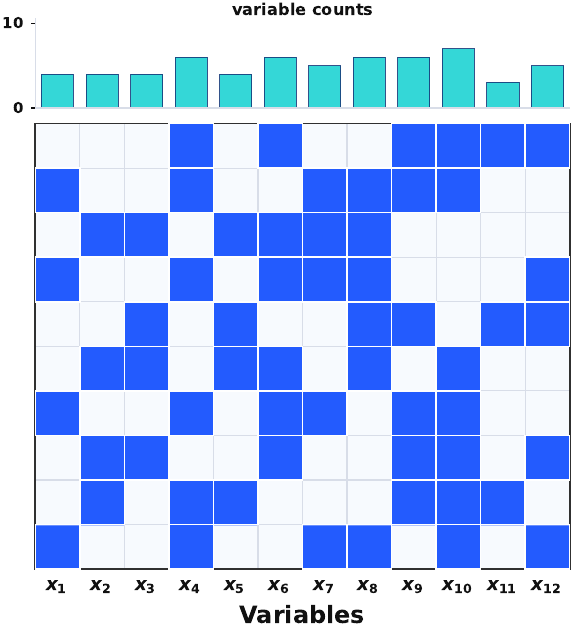}
    \caption{\protect\boldmath
    Medium: $r_{\mathrm{inv}}=0.50$\\[2pt]
    density $=0.50$}
\end{subfigure}
&
\begin{subfigure}[t]{0.315\linewidth}
    \centering
    \includegraphics[width=\linewidth]{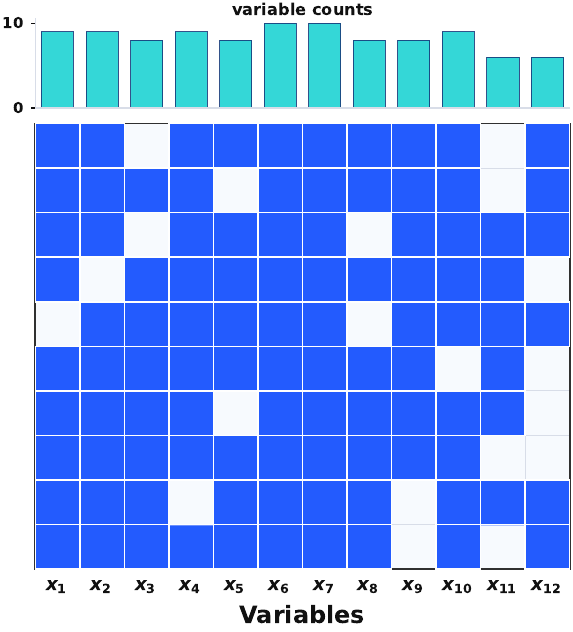}
    \caption{\protect\boldmath
    Dense: $r_{\mathrm{inv}}=0.85$\\[2pt]
    density $=0.83$}
\end{subfigure}

\end{tabular}

\vspace{-3pt}
\caption{Variable-participation patterns induced by different
involvement rates. Filled cells indicate that variable \(x_i\)
participates in constraint \(g_j\), while the marginal bars summarize
participation across variables.}
\label{fig:variable-participation}
\vspace{-8pt}
\end{figure}

Boundary shape describes how an inequality is formed, but not how broadly it
acts across the decision vector. Figure~\ref{fig:variable-participation}
illustrates this effect for a \(12\)-dimensional problem by varying the
involvement rate from sparse to dense. As more variables participate in each
constraint, a coordinate change can influence a larger part of the
constraint set, producing stronger coupling between search directions.
Variable participation thus introduces a distinct source of difficulty:
even a simple boundary can become globally coupled when it depends on most
decision variables
~\cite{hellwig2019benchmarking,gandomi2023gnbgtestsuite}.

\begin{figure}[h]
\centering

\captionsetup[subfigure]{
    font={bf,scriptsize},
    labelfont=bf,
    justification=centering,
    singlelinecheck=false,
    skip=2pt
}

\setlength{\tabcolsep}{2pt}
\renewcommand{\arraystretch}{0.96}

\begin{tabular}{@{}cc@{}}

\begin{subfigure}[t]{0.485\linewidth}
    \centering
    \includegraphics[width=\linewidth]{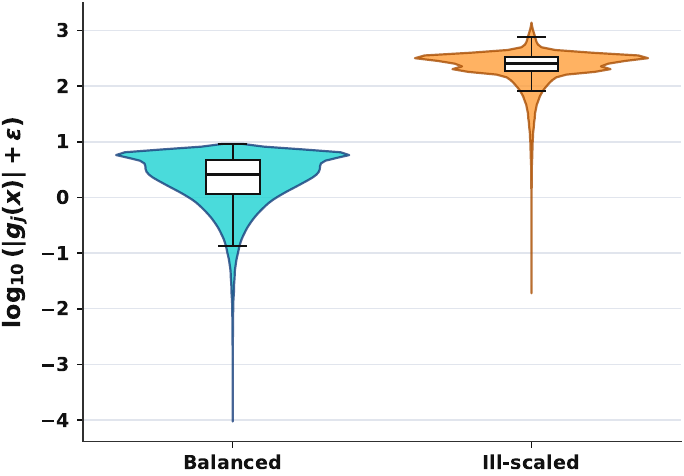}
    \caption{\protect\boldmath Constraint-value scale}
\end{subfigure}
&
\begin{subfigure}[t]{0.485\linewidth}
    \centering
    \includegraphics[width=\linewidth]{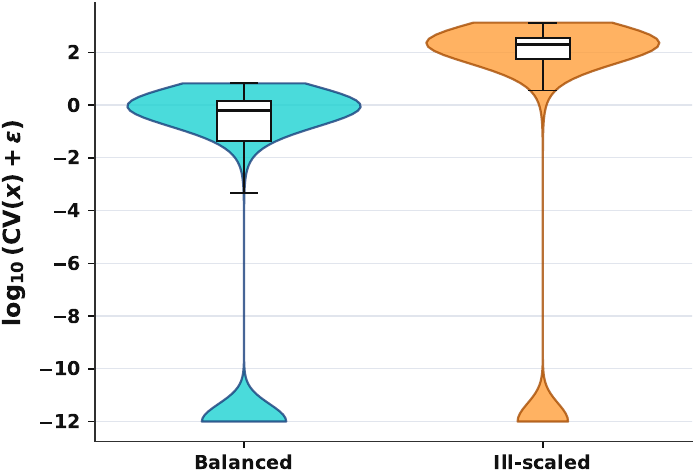}
    \caption{\protect\boldmath Aggregate violation scale}
\end{subfigure}

\end{tabular}

\vspace{-3pt}
\caption{Numerical-scaling diagnostics under balanced and ill-scaled
coefficient regimes. Panel (a) reports
\(\log_{10}(|g_j(x)|+\varepsilon)\), whereas panel (b) reports
\(\log_{10}(\mathrm{CV}(x)+\varepsilon)\). Here,
\(\varepsilon=10^{-12}\) is introduced only to avoid taking the
logarithm of zero and does not affect feasibility classification.}
\label{fig:scaling-diagnostics}
\vspace{-8pt}
\end{figure}

The final comparison changes the numerical scale of the constraint
coefficients. Figure~\ref{fig:scaling-diagnostics} contrasts balanced and
ill-scaled settings using the distributions of individual constraint values
and aggregate constraint violation. In the balanced setting, these values
remain concentrated within a relatively narrow numerical range, so different
constraints contribute on comparable scales. In the ill-scaled setting, the
distributions become much more dispersed and extend across several orders of
magnitude, meaning that a small number of large-valued constraints can dominate
the violation signal. This difference is visible in both panels: the first
shows the wider spread of \(\log_{10}(|g_j(x)|+\varepsilon)\), while the second
shows the corresponding expansion in
\(\log_{10}(\mathrm{CV}(x)+\varepsilon)\). Such scaling changes the numerical
information supplied to penalty methods, stochastic ranking, feasibility
rules, and violation-based tie breaking
~\cite{coello2002theoretical,runarsson2000stochastic,runarsson2005search,deb2000efficient}.

Taken together, these plots provide a progressive view of the GNBG-C
constraint layer. They show how the generator can alter the amount of
feasible space, the local activity of boundaries at the known optimum, the
geometry of the feasibility interface, the degree of variable coupling, the
scale of violation information, and the presence of redundant inequalities.
By isolating these effects one at a time, the visual diagnostics establish
the problem-side interpretation needed for the optimizer study that follows.

\subsection{Baseline evaluation of generated regimes}
\label{subsec:baseline-optimizer-behavior}

The structural diagnostics above establish how the GNBG-C controls are
realized in the generated constraint layer. The next step is to determine
whether these controlled differences remain visible when the resulting
problems are approached as black-box optimization tasks. The experiments in
this subsection therefore examine two complementary stages of constrained
search: entry into the feasible region and improvement of the best feasible
solution after feasibility has been attained. The aim is not to produce a
definitive ranking of optimizers, but to use their responses as evidence that
the generated regimes induce distinguishable search behaviour
~\cite{hansen2021coco,hellwig2019benchmarking,deb2000efficient}.

To examine these effects under a common experimental setting, the baseline
study uses \(d=10\), three GNBG components, and eight feasibility-side
regimes. For each regime, 15 independent problem instances are generated,
and each algorithm is executed for 30 independent runs on every instance,
yielding \(15\times30=450\) runs per algorithm--regime pair. The same
instances are used across all algorithms to ensure a matched comparison.
Each run uses a population size of 100 and a fixed budget of
\(1000d=10{,}000\) evaluations. With eight algorithms and eight regimes, the
study comprises \(8\times8\times15\times30=28{,}800\) optimizer runs.
Objective-, constraint-, and algorithm-run seeds are generated
deterministically from the base seed \(s_0=20260505\), with distinct run
seeds assigned to the 30 repetitions of each algorithm--instance pair. The
experimental protocol is summarized in
Table~\ref{tab:baseline-protocol}.

\begin{table}[h]
\centering
\caption{Experimental protocol for the regime-wise optimizer study.}
\label{tab:baseline-protocol}
\footnotesize
\setlength{\tabcolsep}{3pt}
\renewcommand{\arraystretch}{1.08}

\begin{tabularx}{\columnwidth}{
    @{}
    >{\raggedright\arraybackslash}p{0.58\columnwidth}
    >{\raggedright\arraybackslash}X
    @{}
}
\toprule
Item & Value \\
\midrule
Dimension and GNBG components
& \(d=10,\ o=3\) \\

Feasibility-side regimes
& 8 \\



Instances per regime
& 15 \\

Runs per algorithm--instance
& 30 \\

Runs per algorithm--regime
& 450 \\

Algorithms
& 8 \\

Total optimizer runs
& 28,800 \\

Population size
& 100 \\

Evaluation budget
& \(10{,}000\) per run \\

Checkpoint budgets
& \(2{,}500,\ 5{,}000,\ 10{,}000\) \\

Base seed
& \(20260505\) \\

Feasibility tolerance
& \(10^{-10}\) \\

Static penalty (DE-PEN)
& \(10^6\) \\
\bottomrule
\end{tabularx}
\end{table}

With the problem-generation conditions fixed, the optimizer panel is chosen
to represent several common approaches to constrained population-based
search~\cite{coello2002theoretical,blank2020pymoo}. The three DE variants
share the same underlying search family~\cite{storn1997differential}, but
differ in how feasibility information enters selection. DE-FF applies
feasibility-first constrained comparison~\cite{deb2000efficient}, DE-EPS
uses an adaptive \(\epsilon\)-constraint rule~\cite{takahama2006epsilon},
and DE-PEN incorporates constraint violation through a static
penalty~\cite{coello2002theoretical}. PSO-FF, based on particle swarm
optimization~\cite{kennedy1995particle}, and GA-FF, based on a genetic
algorithm~\cite{holland1975adaptation}, provide feasibility-first population
baselines with different search dynamics. SRES employs stochastic
ranking~\cite{runarsson2000stochastic}, whereas ISRES uses its improved
variant~\cite{runarsson2005search}. Finally, SCIPY-DE uses the differential
evolution implementation provided by SciPy~\cite{virtanen2020scipy} with explicit nonlinear-constraint
handling~\cite{storn1997differential,virtanen2020scipy}. This selection is
intended to reveal different sensitivities to feasible-region scarcity,
active boundaries, violation scaling, and redundant constraint information,
rather than to form an exhaustive algorithm comparison. The optimizer
settings are listed in Table~\ref{tab:algorithm-setup}.

\begin{table}[h]
\centering
\caption{Baseline optimizers and constraint-handling strategies.}
\label{tab:algorithm-setup}
\footnotesize
\setlength{\tabcolsep}{3.5pt}
\renewcommand{\arraystretch}{1.08}
\begin{tabularx}{\columnwidth}{@{}lll>{\raggedright\arraybackslash}X@{}}
\toprule
Algorithm & Backend & Base & Constraint handling \\
\midrule
DE-FF & pymoo~\cite{blank2020pymoo} & DE~\cite{storn1997differential} & feasibility-first constrained comparison~\cite{deb2000efficient} \\
DE-EPS & pymoo & DE & adaptive \(\epsilon\)-constraint handling~\cite{takahama2006epsilon} \\
DE-PEN & pymoo & DE & static penalty~\cite{coello2002theoretical} \\
PSO-FF & pymoo & PSO~\cite{kennedy1995particle} & feasibility-first constrained comparison \\
GA-FF & pymoo & GA~\cite{holland1975adaptation} & feasibility-first constrained comparison \\
SRES & pymoo & SRES~\cite{runarsson2000stochastic} & stochastic ranking \\
ISRES & pymoo & ISRES~\cite{runarsson2005search} & improved stochastic ranking \\
SCIPY-DE & scipy~\cite{virtanen2020scipy} & DE & \texttt{NonlinearConstraint} \\
\bottomrule
\end{tabularx}
\end{table}

Performance is recorded separately for feasibility discovery and feasible
objective improvement. A candidate solution is treated as feasible when
\(\mathrm{CV}(x)\leq10^{-10}\), and the first feasible evaluation is the
earliest evaluation at which this condition is satisfied
~\cite{liang2006problem,deb2000efficient}. Once feasibility has been found,
progress is measured by the best feasible solution encountered so far. If
this solution is denoted by \(x_{\mathrm{bf}}\), and \(f^\star\) is the known
objective value at the reference optimum, the best-feasible gap is
\begin{equation}
    \Delta f_{\mathrm{bf}}=f(x_{\mathrm{bf}})-f^\star.
    \label{eq:diagnostic-best-feasible-gap}
\end{equation}
This quantity excludes objective improvement obtained outside the feasible
region and therefore isolates progress on the constrained problem
~\cite{deb2000efficient,coello2002theoretical}.

Checkpoint gaps are evaluated at
\(B\in\{2{,}500,5{,}000,10{,}000\}\). For each run, the latest recorded
best-so-far state satisfying \(n_{\mathrm{eval}}\leq B\) is used. Means and
standard deviations are calculated over runs that have found a feasible
solution by the corresponding checkpoint, and the associated feasible-run
counts are reported explicitly.

At the final budget, feasible objective quality is also summarized using the
reporting cutoffs \(\tau\in\{25,50\}\). For algorithm \(a\), regime \(c\),
and budget \(B\), the feasible-gap attainment rate is
\begin{equation}
\label{eq:gap-attainment-rate}
\mathrm{AR}_{a,c}(\tau;B)=
\frac{100}{N_{\mathrm{inst}}N_{\mathrm{run}}}
\sum_{i=1}^{N_{\mathrm{inst}}}
\sum_{r=1}^{N_{\mathrm{run}}}
\mathbb{I}\!\left[
\Delta f_{\mathrm{bf}}^{(a,c,i,r)}(B)\leq\tau
\right].
\end{equation}
Here \(N_{\mathrm{inst}}=15\) and \(N_{\mathrm{run}}=30\), so all 450 runs
remain in the denominator. Runs that do not produce a feasible solution are
counted as non-attaining. The two cutoffs are post-hoc reporting criteria,
not strict solution tolerances, and the percentages are rounded to one
decimal place.

The eight regimes specify which feasibility-side property is emphasized in
each group of generated problems. Their effective settings are reported in
Table~\ref{tab:regime-configurations}. The column \emph{Total (P+R)} gives
the total number of constraints and separates the primary constraints
\(P\), which define the feasible region, from the redundant constraints
\(R\); the value of \(R\) therefore also reports the redundancy level.
The \emph{Family mix} lists the numbers of linear, convex-quadratic,
nonconvex-quadratic, polynomial, and transcendental primary constraints,
in that order.

The column \emph{Support / \(r_{\mathrm{inv}}\)} reports both the number of
decision variables involved in each primary constraint and the corresponding
involvement rate. For \(d=10\), a support size of \(2\) gives
\(r_{\mathrm{inv}}=0.20\), whereas the range \(7\)--\(10\) corresponds to
\(r_{\mathrm{inv}}\in[0.70,1.00]\). Under balanced scaling, coefficient
magnitudes are sampled uniformly from \([0.5,2.0]\). Under ill-scaled scaling, coefficient magnitudes are sampled as
\(10^{\xi}\), where \(\xi\sim\mathcal{U}(-4,4)\), and therefore span
\([10^{-4},10^{4}]\). The corresponding transcendental frequencies are
sampled as \(10^{\eta}\), where \(\eta\sim\mathcal{U}(0,2)\), and thus
span \([1,10^{2}]\).

The \emph{Active} column gives the requested number of constraints active
at \(x^\star\). The final two columns report the target feasible-region
ratio and the absolute calibration tolerance. All regimes use the common
Sobol-based calibration sample, seed policy, \(\kappa\) interval, expansion
rule, and iteration limit specified in
Section~\ref{subsec:generation-calibration-diagnostics}; only the target
ratio and calibration tolerance vary across the rows of the table.

\begin{table}[h]
\centering
\caption{Effective configurations of the eight regimes used in the baseline study.}
\label{tab:regime-configurations}
\footnotesize
\setlength{\tabcolsep}{2.2pt}
\renewcommand{\arraystretch}{1.08}
\resizebox{\columnwidth}{!}{%
\begin{tabular}{lccccccc}
\toprule
Regime
& Total (P+R)
& Family mix
& Support / \(r_{\mathrm{inv}}\)
& Scaling
& Active
& Target \(\rho\)
& Calib. tol. \\
\midrule
Loose
& 10 (9+1) & 2/2/2/2/1 & 2 / 0.20
& Balanced & 1 & 0.300 & 0.050 \\

Medium
& 10 (9+1) & 2/2/2/2/1 & 2 / 0.20
& Balanced & 2 & 0.050 & 0.025 \\

Strict
& 10 (9+1) & 2/2/2/2/1 & 7--10 / 0.70--1.00
& Balanced & 3 & 0.008 & 0.004 \\

Active-high
& 10 (10+0) & 2/2/2/2/2 & 2 / 0.20
& Balanced & 5 & 0.005 & 0.002 \\

Mostly nonlinear
& 10 (9+1) & 1/1/2/3/2 & 2 / 0.20
& Balanced & 2 & 0.020 & 0.010 \\

Dense
& 12 (11+1) & 3/2/2/2/2 & 7--10 / 0.70--1.00
& Balanced & 3 & 0.002 & 0.002 \\

Ill-scaled
& 12 (11+1) & 3/2/2/2/2 & 7--10 / 0.70--1.00
& Ill-scaled & 3 & 0.030 & 0.015 \\

High-redundancy
& 12 (8+4) & 2/2/2/1/1 & 2 / 0.20
& Balanced & 2 & 0.050 & 0.025 \\
\bottomrule
\end{tabular}
}

\end{table}

All regime tables use the same reporting convention. Gap and first-feasible
entries are mean (standard deviation), feasible-run counts follow the
checkpoint order defined above, AR25 and AR50 retain all 450 runs in the
denominator, and boldface marks the best value in each column.

\subsubsection{Loose regime}
\label{subsubsec:loose-checkpoint-results}

The loose regime provides the least restrictive target-controlled setting. It uses sparse two-variable supports, one active constraint at the known optimum, and a target feasibility ratio of \(0.30\). The 15 instances achieved \(\hat{\rho}=0.2384\,(0.1029)\); the calibration target was reached for 12 instances, and all instances were valid. Results are reported in Table~\ref{tab:loose-checkpoint-results}.

\begin{table}[h]
\centering
\caption{Optimizer results for the loose regime.}
\label{tab:loose-checkpoint-results}
\footnotesize
\setlength{\tabcolsep}{2.2pt}
\renewcommand{\arraystretch}{1.04}
\resizebox{\columnwidth}{!}{%
\begin{tabular}{lccccccc}
\toprule
Algorithm
& Gap at 2,500 FE
& Gap at 5,000 FE
& Gap at 10,000 FE
& Feasible runs
& First feasible FE
& AR25 (\%)
& AR50 (\%) \\
\midrule
DE-FF & 89.30 (12.58) & 67.72 (14.71) & 49.13 (17.84) & 450/450/450 & \textbf{8.34 (17.08)} & 3.6 & 58.7 \\
DE-EPS & 89.23 (12.96) & 67.76 (14.08) & 47.26 (17.32) & 450/450/450 & 8.58 (17.65) & 2.7 & 64.7 \\
DE-PEN & 86.46 (13.03) & 66.18 (14.37) & 48.31 (17.77) & 450/450/450 & 9.52 (20.59) & 4.9 & 61.6 \\
PSO-FF & \textbf{63.24 (24.94)} & 52.04 (28.68) & 45.69 (31.79) & 450/450/450 & 9.91 (22.08) & 33.3 & 58.4 \\
GA-FF & 66.43 (15.74) & 49.56 (20.75) & 43.94 (23.50) & 450/450/450 & 10.93 (30.38) & 27.3 & 59.6 \\
SRES & 78.01 (13.84) & \textbf{45.04 (17.16)} & \textbf{22.11 (21.84)} & 450/450/450 & 10.32 (24.25) & \textbf{73.6} & \textbf{80.2} \\
ISRES & 92.37 (12.52) & 64.86 (15.72) & 34.07 (20.80) & 450/450/450 & 11.05 (26.12) & 51.3 & 74.4 \\
SCIPY-DE & 78.32 (15.41) & 51.94 (20.54) & 30.80 (25.74) & 450/450/450 & 9.62 (21.60) & 62.9 & 73.6 \\
\bottomrule
\end{tabular}
}

\end{table}

All methods had identified a feasible solution in every run by the first checkpoint, so the main separation arose after feasibility. PSO-FF obtained the smallest mean gap at \(2{,}500\) evaluations, whereas SRES obtained the smallest gaps at \(5{,}000\) and \(10{,}000\) evaluations and the highest attainment rates. The earliest mean feasible entry, obtained by DE-FF, therefore did not imply the best final feasible objective quality.

\subsubsection{Medium regime}
\label{subsubsec:medium-checkpoint-results}

The medium regime retains the sparse support pattern but reduces the target feasibility ratio to \(0.05\) and places two constraints active at the known optimum. The achieved ratio was \(0.0440\,(0.0096)\), and all 15 instances reached the calibration target and passed the validity checks. Results are reported in Table~\ref{tab:medium-checkpoint-results}.

\begin{table}[h]
\centering
\caption{Optimizer results for the medium regime.}
\label{tab:medium-checkpoint-results}
\footnotesize
\setlength{\tabcolsep}{2.2pt}
\renewcommand{\arraystretch}{1.04}
\resizebox{\columnwidth}{!}{%
\begin{tabular}{lccccccc}
\toprule
Algorithm
& Gap at 2,500 FE
& Gap at 5,000 FE
& Gap at 10,000 FE
& Feasible runs
& First feasible FE
& AR25 (\%)
& AR50 (\%) \\
\midrule
DE-FF & 89.51 (15.52) & 66.98 (15.60) & 47.67 (17.53) & 450/450/450 & \textbf{22.21 (20.49)} & 4.4 & 70.2 \\
DE-EPS & 91.33 (16.97) & 67.53 (17.56) & 46.35 (19.68) & 450/450/450 & 25.89 (28.13) & 4.9 & 73.6 \\
DE-PEN & 87.14 (14.93) & 66.15 (16.19) & 48.09 (19.58) & 450/450/450 & 23.97 (24.01) & 2.2 & 72.9 \\
PSO-FF & 70.89 (30.27) & 58.65 (33.58) & 51.06 (36.46) & 450/450/450 & 23.11 (24.21) & 31.6 & 57.8 \\
GA-FF & \textbf{69.30 (18.59)} & 50.70 (21.43) & 43.93 (23.84) & 450/450/450 & 24.08 (25.36) & 23.6 & 66.0 \\
SRES & 80.43 (15.09) & \textbf{46.75 (16.88)} & \textbf{21.25 (21.17)} & 450/450/450 & 24.73 (26.53) & \textbf{77.1} & \textbf{88.4} \\
ISRES & 93.89 (14.23) & 65.01 (15.06) & 32.76 (18.94) & 450/450/450 & 24.22 (27.47) & 50.2 & 84.2 \\
SCIPY-DE & 84.13 (17.90) & 56.26 (19.57) & 32.03 (24.01) & 450/450/450 & 23.49 (22.93) & 56.4 & 81.8 \\
\bottomrule
\end{tabular}
}

\end{table}

Feasible coverage was again complete by the first checkpoint, although first feasible evaluations were generally later than in the loose regime. GA-FF produced the smallest mean gap at \(2{,}500\) evaluations, while SRES produced the smallest later gaps and the highest attainment rates. The medium regime therefore continued to distinguish post-feasibility improvement without causing persistent feasibility failures.

\subsubsection{Strict regime}
\label{subsubsec:strict-checkpoint-results}

The strict regime combines a target feasibility ratio of \(0.008\) with dense supports and three active constraints at the known optimum. The achieved ratio was \(0.0060\,(0.0043)\); 10 of the 15 instances reached the calibration target, while all instances remained valid and preserved the requested active set. Results are reported in Table~\ref{tab:strict-checkpoint-results}.

\begin{table}[h]
\centering
\caption{Optimizer results for the strict regime.}
\label{tab:strict-checkpoint-results}
\footnotesize
\setlength{\tabcolsep}{2.2pt}
\renewcommand{\arraystretch}{1.04}
\resizebox{\columnwidth}{!}{%
\begin{tabular}{lccccccc}
\toprule
Algorithm
& Gap at 2,500 FE
& Gap at 5,000 FE
& Gap at 10,000 FE
& Feasible runs
& First feasible FE
& AR25 (\%)
& AR50 (\%) \\
\midrule
DE-FF & 106.58 (18.93) & 84.85 (18.85) & 66.55 (20.98) & 450/450/450 & 244.08 (312.32) & 0.0 & 25.8 \\
DE-EPS & 108.91 (21.66) & 81.59 (20.31) & 57.20 (21.58) & 449/450/450 & 307.21 (472.34) & 0.2 & 43.6 \\
DE-PEN & 106.30 (18.78) & 84.73 (19.69) & 66.50 (21.63) & 450/450/450 & 244.52 (310.39) & 0.0 & 27.3 \\
PSO-FF & 93.39 (33.42) & 79.63 (38.03) & 70.91 (42.23) & 450/450/450 & \textbf{160.58 (145.04)} & 20.4 & 39.8 \\
GA-FF & \textbf{92.47 (21.64)} & 72.70 (27.11) & 64.69 (30.58) & 450/450/450 & 211.91 (240.62) & 6.4 & 38.0 \\
SRES & 96.28 (17.67) & \textbf{60.22 (20.29)} & \textbf{30.27 (27.74)} & 450/450/450 & 250.99 (354.97) & \textbf{68.9} & \textbf{79.3} \\
ISRES & 110.64 (18.94) & 78.37 (18.39) & 41.40 (23.28) & 448/450/450 & 289.54 (500.86) & 24.0 & 77.8 \\
SCIPY-DE & 109.95 (25.54) & 83.67 (29.14) & 52.96 (28.28) & 417/444/450 & 485.27 (1056.41) & 14.7 & 56.0 \\
\bottomrule
\end{tabular}
}

\end{table}

The stricter feasible region delayed entry for several methods: at \(2{,}500\) evaluations, SCIPY-DE had feasible solutions in 417 of 450 runs, while DE-EPS and ISRES had 449 and 448, respectively. All methods reached complete coverage by the final checkpoint, but the final gaps remained larger than in the loose and medium regimes. SRES achieved the smallest \(5{,}000\)- and \(10{,}000\)-evaluation gaps and the highest attainment rates.

\begin{figure}[h]
    \centering
    \includegraphics[width=\linewidth]{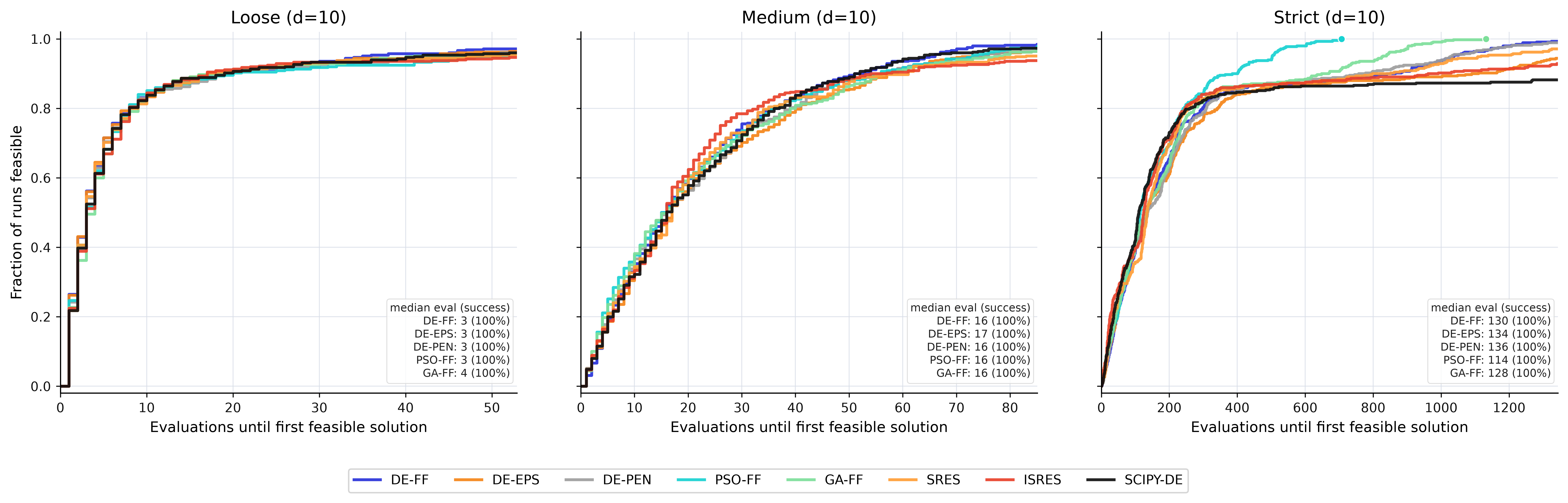}
    \caption{Feasibility discovery in the loose, medium, and strict regimes. Each curve reports the fraction of runs that have identified a feasible solution by the indicated evaluation count.}
    \label{fig:feasibility-discovery}
\end{figure}

The loose, medium, and strict regimes form an ordered target-feasibility
stress sequence rather than a one-factor ablation, because their
active-set and support settings also become more restrictive.
Figure~\ref{fig:feasibility-discovery} collects their feasible-entry
trajectories. The curves shift to later evaluations across the sequence,
while the regime tables show that subsequent feasible objective progress
remains algorithm dependent.

The remaining regimes alter other properties of the constraint layer.
They are considered in the same format so that feasible coverage and
post-feasibility progress remain directly comparable.

\subsubsection{Active-high regime}
\label{subsubsec:active-high-checkpoint-results}

The active-high regime isolates local boundary pressure by placing five constraints active at the known optimum while retaining sparse supports. Its target feasibility ratio is \(0.005\), and the achieved ratio was \(0.0027\,(0.0025)\). Eight instances reached the calibration target; all 15 instances were valid and realized the requested five active constraints. Results are reported in Table~\ref{tab:active-high-checkpoint-results}.

\begin{table}[h]
\centering
\caption{Optimizer results for the active-high regime.}
\label{tab:active-high-checkpoint-results}
\footnotesize
\setlength{\tabcolsep}{2.2pt}
\renewcommand{\arraystretch}{1.04}
\resizebox{\columnwidth}{!}{%
\begin{tabular}{lccccccc}
\toprule
Algorithm
& Gap at 2,500 FE
& Gap at 5,000 FE
& Gap at 10,000 FE
& Feasible runs
& First feasible FE
& AR25 (\%)
& AR50 (\%) \\
\midrule
DE-FF & 108.50 (20.29) & 83.46 (20.16) & 63.65 (21.54) & 450/450/450 & 436.33 (421.51) & 2.4 & 30.4 \\
DE-EPS & 118.53 (27.44) & 92.69 (26.38) & 61.29 (21.10) & 422/449/449 & 683.45 (943.68) & 3.1 & 35.6 \\
DE-PEN & 105.08 (20.22) & 80.42 (20.16) & 62.48 (22.21) & 449/450/450 & 416.89 (405.76) & 3.8 & 33.3 \\
PSO-FF & 96.32 (31.68) & 83.07 (35.83) & 73.75 (39.58) & 433/434/434 & \textbf{268.89 (234.30)} & 12.7 & 33.6 \\
GA-FF & \textbf{92.83 (24.89)} & 75.97 (29.01) & 69.85 (31.32) & 450/450/450 & 388.08 (321.85) & 8.4 & 30.4 \\
SRES & 103.75 (22.24) & \textbf{65.42 (21.70)} & \textbf{40.26 (30.00)} & 446/450/450 & 495.66 (586.27) & \textbf{46.7} & \textbf{67.8} \\
ISRES & 118.93 (24.82) & 85.07 (21.10) & 50.41 (26.21) & 426/450/450 & 595.10 (782.79) & 15.3 & 61.6 \\
SCIPY-DE & 116.46 (30.69) & 93.86 (38.05) & 64.07 (32.81) & 371/431/449 & 1052.76 (1675.81) & 8.2 & 41.1 \\
\bottomrule
\end{tabular}
}

\end{table}

The five active boundaries affected both coverage and feasible improvement. At the final checkpoint, PSO-FF had feasible solutions in 434 of 450 runs, while DE-EPS and SCIPY-DE had 449; the remaining methods reached complete coverage. SRES retained complete coverage and achieved the smallest later gaps and the highest attainment rates, indicating that local activity continued to influence progress after feasibility was found.

\subsubsection{Mostly nonlinear regime}
\label{subsubsec:mostly-nonlinear-checkpoint-results}

The mostly nonlinear regime increases the share of polynomial, transcendental, and nonconvex-quadratic boundaries while retaining sparse supports and two active constraints. The target feasibility ratio is \(0.02\), and the achieved ratio was \(0.0166\,(0.0052)\). Fourteen instances reached the calibration target, and all instances were valid. Results are reported in Table~\ref{tab:mostly-nonlinear-checkpoint-results}.

\begin{table}[h]
\centering
\caption{Optimizer results for the mostly nonlinear regime.}
\label{tab:mostly-nonlinear-checkpoint-results}
\footnotesize
\setlength{\tabcolsep}{2.2pt}
\renewcommand{\arraystretch}{1.04}
\resizebox{\columnwidth}{!}{%
\begin{tabular}{lccccccc}
\toprule
Algorithm
& Gap at 2,500 FE
& Gap at 5,000 FE
& Gap at 10,000 FE
& Feasible runs
& First feasible FE
& AR25 (\%)
& AR50 (\%) \\
\midrule
DE-FF & 97.42 (18.31) & 72.31 (18.17) & 51.45 (18.02) & 450/450/450 & 62.37 (62.73) & 0.2 & 61.3 \\
DE-EPS & 101.22 (20.46) & 73.87 (20.27) & 48.55 (19.47) & 450/450/450 & 65.49 (77.97) & 1.8 & 69.3 \\
DE-PEN & 94.65 (18.44) & 69.84 (18.34) & 50.20 (19.57) & 450/450/450 & 66.66 (68.51) & 2.0 & 62.0 \\
PSO-FF & 82.46 (35.00) & 71.42 (38.32) & 64.67 (41.18) & 450/450/450 & 56.31 (49.54) & 23.6 & 43.6 \\
GA-FF & \textbf{77.65 (21.00)} & 58.09 (25.96) & 51.33 (29.08) & 450/450/450 & 60.37 (64.07) & 20.7 & 56.9 \\
SRES & 87.29 (16.18) & \textbf{49.23 (15.82)} & \textbf{21.17 (19.91)} & 450/450/450 & 57.52 (60.62) & \textbf{82.4} & \textbf{88.4} \\
ISRES & 102.09 (16.67) & 69.57 (16.83) & 33.43 (18.83) & 450/450/450 & 56.70 (56.10) & 41.1 & 84.0 \\
SCIPY-DE & 94.69 (20.46) & 63.06 (21.63) & 34.47 (22.89) & 450/450/450 & \textbf{55.84 (52.83)} & 45.8 & 79.3 \\
\bottomrule
\end{tabular}
}

\end{table}

All methods reached complete feasible coverage by the first checkpoint, so the nonlinear family composition mainly separated their feasible objective progress. GA-FF obtained the smallest mean gap at \(2{,}500\) evaluations, whereas SRES obtained the smallest later gaps and the highest attainment rates. The effect of this regime is therefore expressed primarily through post-feasibility improvement rather than delayed entry.

\subsubsection{Dense regime}
\label{subsubsec:dense-checkpoint-results}

The dense regime increases variable coupling by allowing each primary constraint to involve 7--10 of the 10 decision variables. It uses three active constraints and a target feasibility ratio of \(0.002\). The achieved ratio was \(0.0016\,(0.0009)\), and all 15 instances reached the calibration target and passed the validity checks. Results are reported in Table~\ref{tab:dense-checkpoint-results}.

\begin{table}[h]
\centering
\caption{Optimizer results for the dense regime.}
\label{tab:dense-checkpoint-results}
\footnotesize
\setlength{\tabcolsep}{2.2pt}
\renewcommand{\arraystretch}{1.04}
\resizebox{\columnwidth}{!}{%
\begin{tabular}{lccccccc}
\toprule
Algorithm
& Gap at 2,500 FE
& Gap at 5,000 FE
& Gap at 10,000 FE
& Feasible runs
& First feasible FE
& AR25 (\%)
& AR50 (\%) \\
\midrule
DE-FF & 104.14 (18.25) & 81.30 (18.83) & 61.27 (20.33) & 444/449/450 & 366.17 (571.13) & 0.0 & 30.9 \\
DE-EPS & 105.94 (20.37) & 75.64 (21.67) & 51.89 (21.54) & 447/450/450 & 361.79 (434.65) & 0.0 & 66.4 \\
DE-PEN & 103.39 (19.63) & 80.55 (19.63) & 60.35 (19.98) & 449/450/450 & 333.38 (374.65) & 0.0 & 33.8 \\
PSO-FF & \textbf{90.62 (35.94)} & 76.41 (41.19) & 67.22 (47.59) & 429/434/444 & 376.46 (996.00) & 22.4 & 50.2 \\
GA-FF & 90.75 (22.71) & 69.64 (27.56) & 60.39 (31.42) & 443/443/444 & 300.81 (390.97) & 9.6 & 47.6 \\
SRES & 90.98 (16.52) & \textbf{52.20 (16.02)} & \textbf{17.47 (14.27)} & 450/450/450 & \textbf{283.52 (285.29)} & \textbf{88.0} & \textbf{97.3} \\
ISRES & 106.57 (15.89) & 72.74 (17.92) & 34.43 (19.08) & 449/450/450 & 298.04 (342.17) & 35.6 & 88.7 \\
SCIPY-DE & 113.57 (21.46) & 80.38 (23.86) & 49.98 (31.43) & 420/424/441 & 587.43 (1535.54) & 10.9 & 70.7 \\
\bottomrule
\end{tabular}
}

\end{table}

Dense variable coupling produced a wider separation in feasible coverage. At \(10{,}000\) evaluations, PSO-FF and GA-FF had feasible solutions in 444 of 450 runs and SCIPY-DE in 441, whereas SRES, ISRES, and the three DE variants reached or approached complete coverage. SRES achieved the smallest final checkpoint gap and the highest attainment rates, showing that dense coupling affected both feasible discovery and subsequent improvement.

\subsubsection{Ill-scaled regime}
\label{subsubsec:ill-scaled-checkpoint-results}

The ill-scaled regime retains dense supports but draws coefficient magnitudes over several orders of magnitude. It uses three active constraints and a target feasibility ratio of \(0.03\). The achieved ratio was \(0.0186\,(0.0128)\); nine instances reached the calibration target, and all instances remained valid. Results are reported in Table~\ref{tab:ill-scaled-checkpoint-results}.

\begin{table}[h]
\centering
\caption{Optimizer results for the ill-scaled regime.}
\label{tab:ill-scaled-checkpoint-results}
\footnotesize
\setlength{\tabcolsep}{2.2pt}
\renewcommand{\arraystretch}{1.04}
\resizebox{\columnwidth}{!}{%
\begin{tabular}{lccccccc}
\toprule
Algorithm
& Gap at 2,500 FE
& Gap at 5,000 FE
& Gap at 10,000 FE
& Feasible runs
& First feasible FE
& AR25 (\%)
& AR50 (\%) \\
\midrule
DE-FF & 107.07 (20.58) & 87.08 (23.29) & 65.80 (22.84) & 393/431/450 & 683.05 (1554.64) & 0.0 & 28.7 \\
DE-EPS & 108.96 (22.38) & 86.14 (23.75) & 60.79 (22.52) & 382/406/449 & 901.26 (1971.38) & 0.0 & 39.3 \\
DE-PEN & 106.51 (19.36) & 86.04 (22.05) & 64.36 (20.55) & 398/436/449 & 640.58 (1457.02) & 0.0 & 28.9 \\
PSO-FF & 95.25 (32.79) & 82.83 (36.25) & 74.69 (40.66) & 431/435/435 & 245.88 (508.30) & 12.2 & 30.2 \\
GA-FF & \textbf{89.57 (23.82)} & 74.11 (29.47) & 67.27 (32.25) & 401/432/441 & 507.66 (1223.75) & 9.3 & 35.1 \\
SRES & 100.44 (21.43) & \textbf{65.89 (25.13)} & \textbf{36.76 (27.77)} & 393/445/449 & 572.97 (1216.40) & \textbf{49.8} & \textbf{70.4} \\
ISRES & 111.27 (21.23) & 82.65 (22.81) & 46.53 (24.69) & 387/421/450 & 806.24 (1771.63) & 21.3 & 63.1 \\
SCIPY-DE & 104.67 (23.91) & 76.35 (26.18) & 48.05 (28.37) & 389/390/390 & \textbf{138.04 (349.14)} & 22.7 & 52.4 \\
\bottomrule
\end{tabular}
}

\end{table}

The ill-scaled regime produced the largest dispersion in feasible coverage: final checkpoint counts ranged from 390 to 450 runs. SCIPY-DE had the earliest mean feasible entry among its feasible runs but reached feasibility in only 390 of 450 runs, so entry speed cannot be interpreted independently of coverage. SRES combined near-complete coverage with the smallest final checkpoint gap and the highest attainment rates.

\subsubsection{High-redundancy regime}
\label{subsubsec:high-redundancy-checkpoint-results}

The high-redundancy regime uses eight primary and four redundant constraints, compared with one redundant constraint in the target-controlled regimes. It retains sparse supports, two active constraints, and a target feasibility ratio of \(0.05\). The achieved ratio was \(0.0407\,(0.0174)\); 13 instances reached the calibration target, and all redundancy and validity checks passed. Results are reported in Table~\ref{tab:high-redundancy-checkpoint-results}.

\begin{table}[h]
\centering
\caption{Optimizer results for the high-redundancy regime.}
\label{tab:high-redundancy-checkpoint-results}
\footnotesize
\setlength{\tabcolsep}{2.2pt}
\renewcommand{\arraystretch}{1.04}
\resizebox{\columnwidth}{!}{%
\begin{tabular}{lccccccc}
\toprule
Algorithm
& Gap at 2,500 FE
& Gap at 5,000 FE
& Gap at 10,000 FE
& Feasible runs
& First feasible FE
& AR25 (\%)
& AR50 (\%) \\
\midrule
DE-FF & 98.44 (14.88) & 74.40 (17.27) & 52.83 (21.00) & 450/450/450 & 35.42 (55.56) & 0.4 & 63.6 \\
DE-EPS & 100.99 (16.13) & 75.44 (18.14) & 49.41 (20.60) & 450/450/450 & 34.32 (46.19) & 3.1 & 66.2 \\
DE-PEN & 95.17 (14.73) & 72.07 (16.62) & 51.32 (20.56) & 450/450/450 & 35.37 (54.12) & 1.6 & 63.3 \\
PSO-FF & 80.97 (32.71) & 65.94 (38.28) & 56.33 (42.98) & 450/450/450 & 33.93 (48.78) & 35.8 & 52.9 \\
GA-FF & \textbf{77.23 (18.82)} & 55.47 (25.56) & 47.00 (29.34) & 450/450/450 & 38.26 (60.53) & 31.3 & 61.8 \\
SRES & 88.55 (14.77) & \textbf{51.63 (18.57)} & \textbf{22.47 (24.78)} & 450/450/450 & 37.42 (63.07) & \textbf{82.9} & \textbf{85.6} \\
ISRES & 101.09 (13.16) & 70.60 (16.72) & 34.06 (22.60) & 450/450/450 & \textbf{32.95 (47.96)} & 48.2 & 83.6 \\
SCIPY-DE & 92.98 (16.06) & 62.75 (20.03) & 34.62 (25.96) & 450/450/450 & 35.19 (58.12) & 53.8 & 79.6 \\
\bottomrule
\end{tabular}
}

\end{table}

\begin{figure*}[h]
    \centering
    \includegraphics[width=\linewidth]{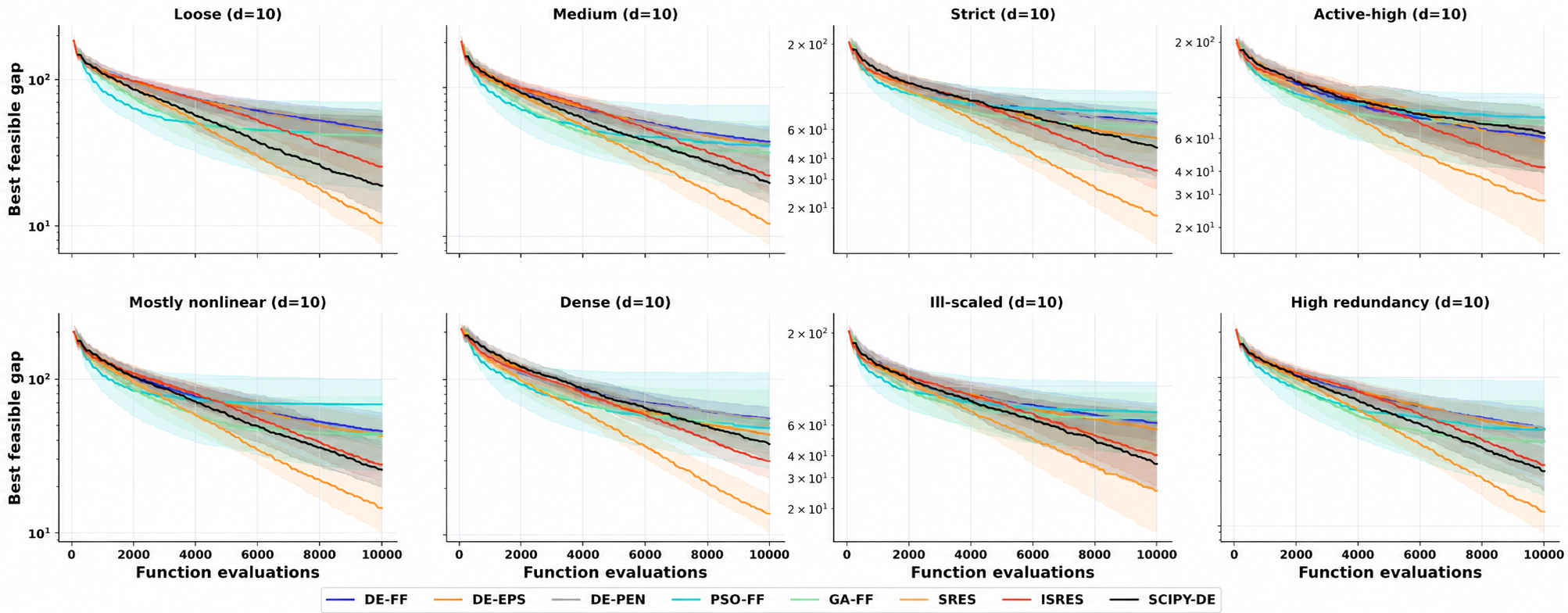}
    \caption{Best-feasible-gap convergence across the eight generated
    regimes. The curves summarize feasible objective improvement over the
    evaluation budget; lower values indicate better progress.}
    \label{fig:best-feasible-gap-convergence}
\end{figure*}

All methods identified feasible solutions in every run by the first checkpoint. GA-FF obtained the smallest mean gap at \(2{,}500\) evaluations, while SRES obtained the smallest later gaps and the highest attainment rates; ISRES recorded the earliest mean feasible entry. In this regime, the main algorithmic differences therefore appeared after feasibility rather than in feasible discovery.

The regime tables provide fixed-budget snapshots; the convergence view
in Figure~\ref{fig:best-feasible-gap-convergence} shows how those
snapshots are reached. Loose, medium, mostly nonlinear, and
high-redundancy instances generally attain feasibility early, so their
separation is dominated by subsequent feasible objective improvement.
Strict, active-high, dense, and ill-scaled instances produce stronger
separation in feasible coverage or later progress. Across these cases,
several methods continue improving throughout the second half of the
budget, while others retain comparatively large gaps.

\begin{figure}[h]
    \centering
    \includegraphics[width=0.8\linewidth]{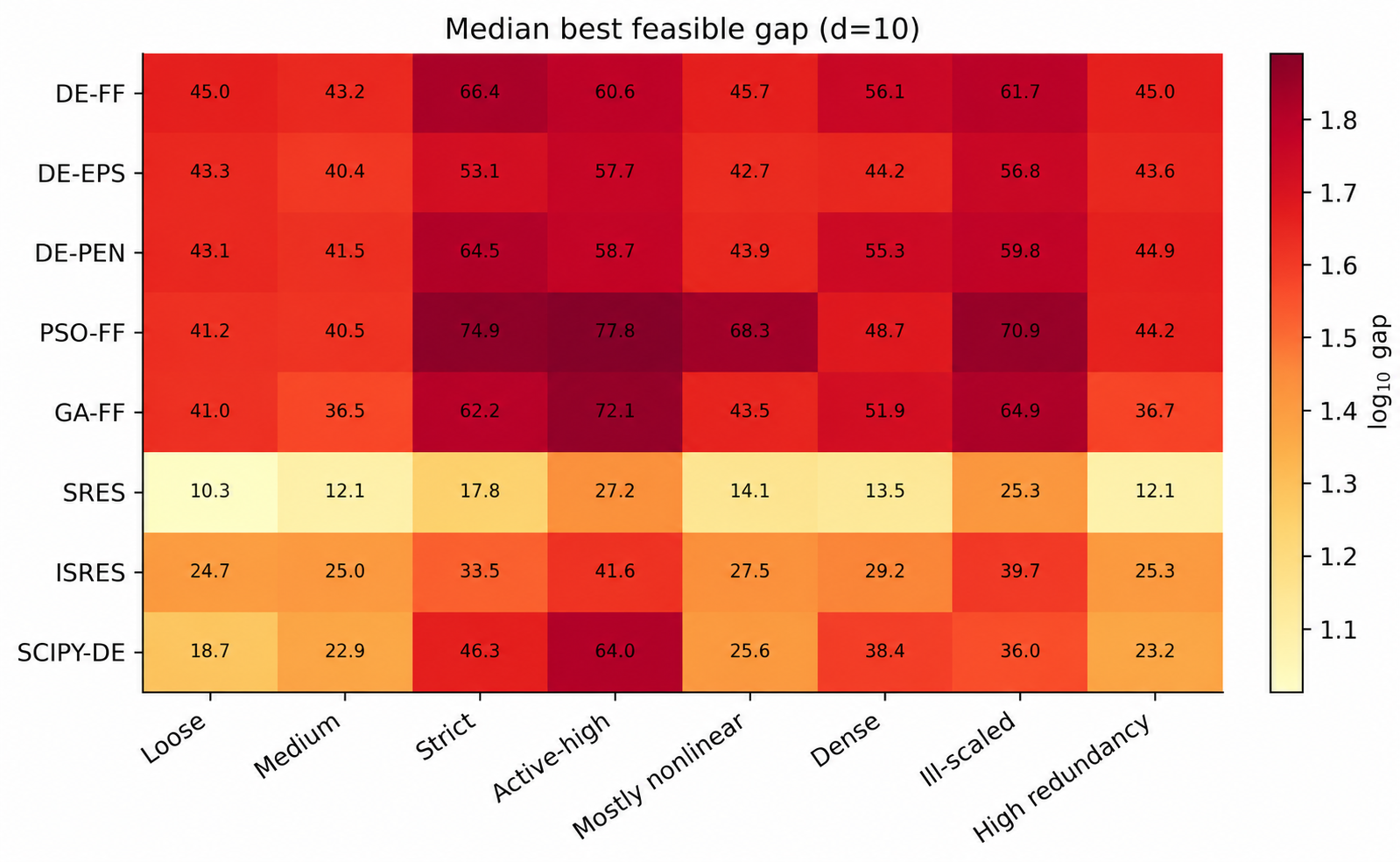}
    \caption{Final median best-feasible gap across algorithms and generated
    regimes. Cell labels report the raw median best-feasible gaps, whereas
    the cell colours encode the logarithmically transformed values,
    \(\log_{10}\!\left(
    \max\!\left\{
    \widetilde{\Delta f}_{\mathrm{bf}},\,10^{-12}
    \right\}
    \right)\),
    where \(\widetilde{\Delta f}_{\mathrm{bf}}\) denotes the median
    best-feasible gap for the corresponding algorithm--regime pair.
    The numerical floor of \(10^{-12}\) is applied only to enable zero
    values to be displayed on the logarithmic colour scale. Lower raw
    values indicate better feasible objective quality at the end of the
    evaluation budget.}
    \label{fig:final-performance-summary}
\end{figure}

Figure~\ref{fig:final-performance-summary} compresses the endpoint
results into a single algorithm--regime view. The raw labels preserve
the final median gaps, while logarithmic colouring makes differences
across the full numerical range visible. The pattern is not a single
ordering of either algorithms or regimes: methods that enter feasibility
quickly do not always obtain the best final feasible objective value,
and the effect of a regime depends on the constraint-handling mechanism.

The realized benchmark diagnostics are summarized in
Table~\ref{tab:structural-diagnostics}. All 120 instances preserved
feasibility of the known optimum and realized the requested active-set
size; no weak active constraints were selected, and all redundancy
checks passed. The achieved feasibility ratios follow the intended
regimes, but some instances did not reach the requested ratio because
zero-slack active constraints are unaffected by the shared calibration
multiplier.

\begin{table}[h]
\centering
\caption{Realized feasibility-ratio and active-set diagnostics for the
\(d=10\) regimes; achieved ratios are mean (standard deviation) over
15 instances.}
\label{tab:structural-diagnostics}
\footnotesize
\setlength{\tabcolsep}{2.4pt}
\renewcommand{\arraystretch}{1.08}
\resizebox{\columnwidth}{!}{%
\begin{tabular}{lccccc}
\toprule
Regime
& Target \(\rho\)
& Achieved \(\hat{\rho}\)
& Active req./real.
& Target reached
& Valid \\
\midrule
Loose
& 0.300 & 0.2384 (0.1029) & 1/1.0 & 12/15 & 15/15 \\

Medium
& 0.050 & 0.0440 (0.0096) & 2/2.0 & 15/15 & 15/15 \\

Strict
& 0.008 & 0.0060 (0.0043) & 3/3.0 & 10/15 & 15/15 \\

Active-high
& 0.005 & 0.0027 (0.0025) & 5/5.0 & 8/15 & 15/15 \\

Mostly nonlinear
& 0.020 & 0.0166 (0.0052) & 2/2.0 & 14/15 & 15/15 \\

Dense
& 0.002 & 0.0016 (0.0009) & 3/3.0 & 15/15 & 15/15 \\

Ill-scaled
& 0.030 & 0.0186 (0.0128) & 3/3.0 & 9/15 & 15/15 \\

High-redundancy
& 0.050 & 0.0407 (0.0174) & 2/2.0 & 13/15 & 15/15 \\
\bottomrule
\end{tabular}
}
\end{table}

Taken together, the results show that the generated regimes expose
distinct phases and sources of constrained-search difficulty. The
feasibility-ratio regimes primarily control the cost of feasible entry;
active, dense, and ill-scaled settings additionally affect feasible
coverage and later improvement; nonlinear boundaries and redundancy
mainly separate methods after feasibility is available. The purpose of
this evaluation is therefore diagnostic: performance differences can be
interpreted through controlled feasibility-side properties rather than
through final objective values alone
~\cite{hansen2021coco,hellwig2019benchmarking}.

\section{Conclusion}
\label{sec:conclusion}

Reliable benchmarking in constrained continuous optimization requires more than a collection of test functions with fixed constraint sets. In constrained search, the objective landscape and the feasible region jointly determine the behavior of an optimizer, yet these two sources of difficulty are often entangled. Poor performance may reflect weak objective-space exploration, delayed feasibility discovery, misleading violation information, unstable progress near restrictive boundaries, or interactions among these factors. Without controlled instance generation, it is difficult to attribute such behavior to a specific source of constrained-search difficulty.

The proposed framework addresses this difficulty by separating feasible-region construction from objective-landscape construction. By anchoring the constraint layer at the known GNBG reference optimum, the framework preserves feasibility while allowing the surrounding admissible region to be shaped systematically. This changes the role of constraints in benchmark generation: they are not merely fixed side conditions attached to a test function, but structured components through which feasibility-side difficulty can be isolated, controlled, and interpreted.

The experimental results support this view. The generated regimes do not simply form a single easy-to-hard ordering; instead, they expose distinct patterns of constrained-search behavior. Loose and medium settings allow feasibility to be reached relatively early, whereas stricter and active-boundary settings place greater pressure on feasible entry and boundary progress. Dense, ill-scaled, and high-redundancy regimes further show that constraint-induced difficulty may arise from variable coupling, violation scaling, or redundant information rather than from feasible-region size alone. In this sense, the proposed framework complements real-world constrained benchmarks: application-driven problems remain essential, but a configurable generator provides the experimental control needed to isolate, combine, and reproduce feasibility-related mechanisms.


A direct extension of the present framework is \textbf{GNBG-C-S} (shifted constrained optimum), in which the original GNBG optimum is deliberately made infeasible and the resulting constrained optimum is separately constructed, approximated, or certified. This mode would broaden the scope of GNBG-C beyond the known-optimum settings considered in the present study, but it requires a reliable procedure for establishing the location and quality of the shifted constrained optimum. Further directions include multi-objective constrained optimization, equality constraints, mixed discrete-continuous variables, dynamic or noisy constraints, and application-informed constraint structures. These extensions would preserve the central goal of the present framework: making constraint-induced difficulty explicit, configurable, and reproducible for systematic optimization benchmarking.

\section*{Contribution Statement}

A.Gh., E.B., A.K., and P.G. contributed equally as first authors. Their contributions included Conceptualization, Methodology, Software, Formal analysis, Investigation, Writing -- original draft, and Writing -- review and editing.

A.G. contributed to Supervision, Conceptualization, Project administration, and Writing -- review and editing. A.G. also served as the corresponding author.

Y.-H.K. and S.M. contributed to Writing -- review and editing of the revised manuscript.

\section*{Conflict of Interest Statement}
The authors have no conflict of interests to declare.

\section*{Funding Statement}
No funding was used to conduct this study.

\section*{Code and Data Availability}

The complete GNBG-C implementation, including the eight regime
definitions, deterministic seed-generation procedures, structural
diagnostics, and scripts for generating benchmark instances and
reproducing the reported figures and tables, is publicly available at
\url{https://github.com/Alighasemzadeh83/GNBG-C}.

The repository also contains the GNBG-C-24 suite, comprising 24 released
instances with their configurations, seeds, known optima, and diagnostic
metadata. The effective configurations used in the baseline study are
summarized in Table~\ref{tab:regime-configurations}. Machine-readable
versions of the resolved regime configurations and the shared calibration
and seed-generation settings are provided in
\path{experiments/paper1/01_effective_regime_configurations.csv}
and
\path{experiments/paper1/02_shared_calibration_settings.json},
respectively. Detailed reproduction instructions are provided in
\path{experiments/paper1/README.md}.

The complete baseline study can be regenerated from the repository root by
executing
\path{python scripts/run_baseline_study.py --mode paper1}.
The generated instances, optimizer outputs, figures, and tables are written
to \path{outputs/paper1/}.

\section*{Acknowledgment}
The authors would like to acknowledge the use of ChatGPT 5.5 solely for refining the language and improving the clarity of the manuscript.

\FloatBarrier
\bibliographystyle{elsarticle-num}
\bibliography{references}

\end{document}